\documentclass[times,final]{article}
    \makeatletter
    \def\ps@pprintTitle{%
    \let\@oddhead\@empty
    \let\@evenhead\@empty
    \def\@oddfoot{}%
    \let\@evenfoot\@oddfoot}
    \makeatother
    \usepackage{lineno}
    \usepackage[breaklinks=true,pdftex]{hyperref}
    \modulolinenumbers[1]
    \usepackage{amsmath,bm}
    \usepackage{svg}
    \usepackage{comment}
    \usepackage{multirow}
    \usepackage[a4paper,
            bindingoffset=0.1in,
            left=1.5cm,
            right=1.5cm,
            top=1.5cm,
            bottom=1.5cm]{geometry}
    \usepackage{ulem}
    \usepackage{booktabs}
    \usepackage{dirtytalk}
    \usepackage{lineno}
    \usepackage{caption}
    \usepackage{amssymb}
    \usepackage{amsmath}
    \usepackage{savesym}
        \usepackage{algorithmic}
    \usepackage{algorithm}
    \savesymbol{AND}
    \usepackage{adjustbox}

    \usepackage{rotating}
    \usepackage{graphicx}
    \usepackage{wrapfig}
    \usepackage{amssymb}
    \usepackage{soul,color}
    \usepackage{subcaption}
    \usepackage{lineno}
\usepackage{xcolor}
\usepackage{url}
    \soulregister\cite7
    \soulregister\ref7
    \soulregister\pageref7

    \usepackage{todonotes}

    \DeclareRobustCommand{\uvec}[1]{{%
    \ifcsname uvec#1\endcsname
    \csname uvec#1\endcsname
    \else
    \bm{\hat{\mathbf{#1}}}%
    \fi}}
\mathchardef\breakingcomma\mathcode`\,
{\catcode`,=\active
  \gdef,{\breakingcomma\discretionary{}{}{}}
}
\newcommand{\mathlist}[1]{$\mathcode`\,=\string"8000 #1$}

\begin{document}

\begin{center}
{\LARGE \textbf{ShipHullGAN: A generic parametric modeller for ship hull design using deep convolutional generative model}}\\
\vspace{0.5cm}
{\small Shahroz Khan$^{1,2,*}$\let\thefootnote\relax\footnote{$^*$Corresponding author. E-mail address: shahroz.khan@strath.ac.uk; shahroz.khan@berkeley.edu (S. Khan)}},
{\small Kosa Goucher-Lambert$^2$},
{\small Konstantinos Kostas$^3$},
{\small Panagiotis Kaklis$^1$}
\\\vspace{0.2cm}
{\small $^1$Department of Naval Architecture, Ocean and Marine Engineering, University of Strathclyde, Glasgow, United Kingdom}\\
{\small $^2$Department of Mechanical Engineering, University of California, Berkeley, United States}\\
{\small $^3$Department of Mechanical and Aerospace Engineering, Nazarbayev University, Astana, Kazakhstan}
\end{center}


\section*{\centering Abstract}
   \begin{figure}[htb!]
    \centering
    \includegraphics[width=01\textwidth]{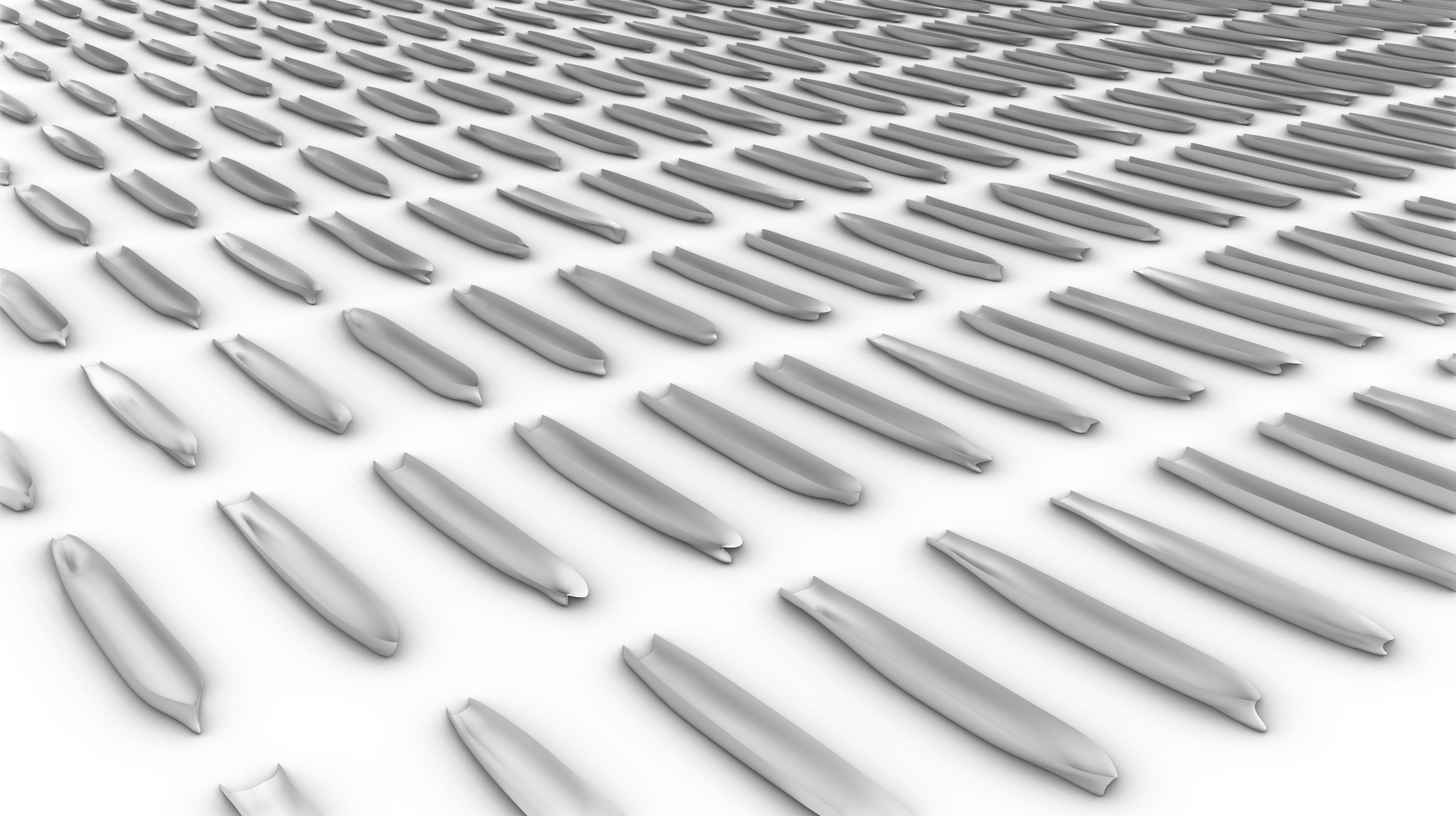}
    \caption{The generic capability of the ShipHullGAN model enables the creation of parametric design variations for a wide range of ship hulls, including both traditional and unconventional forms.} 
    \end{figure}
In this work, we introduce ShipHullGAN, a generic parametric modeller built using deep convolutional generative adversarial networks (GANs) for the versatile representation and generation of ship hulls. At a high level, the new model intends to address the current conservatism in the parametric ship design paradigm, where parametric modellers can only handle a particular ship type. We trained ShipHullGAN on a large dataset of 52,591 \textit{physically validated} designs from a wide range of existing ship types, including container ships, tankers, bulk carriers, tugboats, and crew supply vessels. We developed a new shape extraction and representation strategy to convert all training designs into a common geometric representation of the same resolution, as typically GANs can only accept vectors of fixed dimension as input. A space-filling layer is placed right after the generator component to ensure that the trained generator can cover all design classes. During training, designs are provided in the form of a shape-signature tensor (SST) which harnesses the compact geometric representation using geometric moments that further enable the inexpensive incorporation of physics-informed elements in ship design. We have shown through extensive comparative studies and optimisation cases that ShipHullGAN can generate designs with augmented features resulting in versatile design spaces that produce traditional and novel designs with geometrically valid and practically feasible shapes.
\vspace{0.2cm}\\
\noindent Video abstract: \url{https://youtu.be/LT9Z52vBgzI}
\vspace{0.2cm}\\
\textit{Keywords:} Generative Adversarial Network, Computer-Aided Design, Parametric Design, Geometric Moments, Ship Design, Shape Optimisation



    \section{Introduction}
    \label{intro}
    Recently, machine learning, particularly in the form of scientific machine learning (SciML), has become increasingly prevalent in engineering design. This, on several occasions, has lightened the computational load from traditional solvers by building efficient low or high-fidelity surrogate models that predict performance almost instantly, thus accelerating the simulation-driven design (SDD) process. Although the efforts of integrating SciML in ship design are increasing, the pace is relatively slow compared to other engineering fields.

    Furthermore, there are few efforts to introduce these tools at the preliminary ship design stage, where naval architects and/or involved designers typically identify designs from existing databases while attempting to match new requirements. Afterwards, they may construct a parametric model using a suitable ship-hull surface representation, typically comprising NURBS surface patches or simpler panel meshes. This usually results in a narrow design space permitting only slight variations of a baseline design \cite{nowacki2010five,khan2017customer}. Designers also get inspiration from existing designs while using their features and components to create a small set of potential alternatives. However, embedding these features is a complicated task and constructing a new parametric description for the unique shape using existing strategies is highly expertise-driven and time-intensive. While the current approach to ship design has proven effective for well-established ship types, there may be a need for more radical design ideas in certain situations. This could be when dealing with uncommon requirements that call for an exploration of a richer design space or when revolutionising and redesigning existing ship types, that is likely to arise as a result of major regulation changes, e.g., the IMO 2020 - mandated reduction of emissions, or the emergence of new disrupting technologies in the context of Industry 4.0, such as taking on board non-fossil fuels (ammonia, hydrogen), design and operation of autonomous vessels, etc.; \cite{DKaklisEtAl2023SOME}. Such a strategy will obviously benefit novel design tasks, e.g., \textit{special purpose vessels}, but it can also offer a competitive advantage for traditional players in the industry.

\begin{figure}[htb!]
    \centering
    \includegraphics[width=01\textwidth]{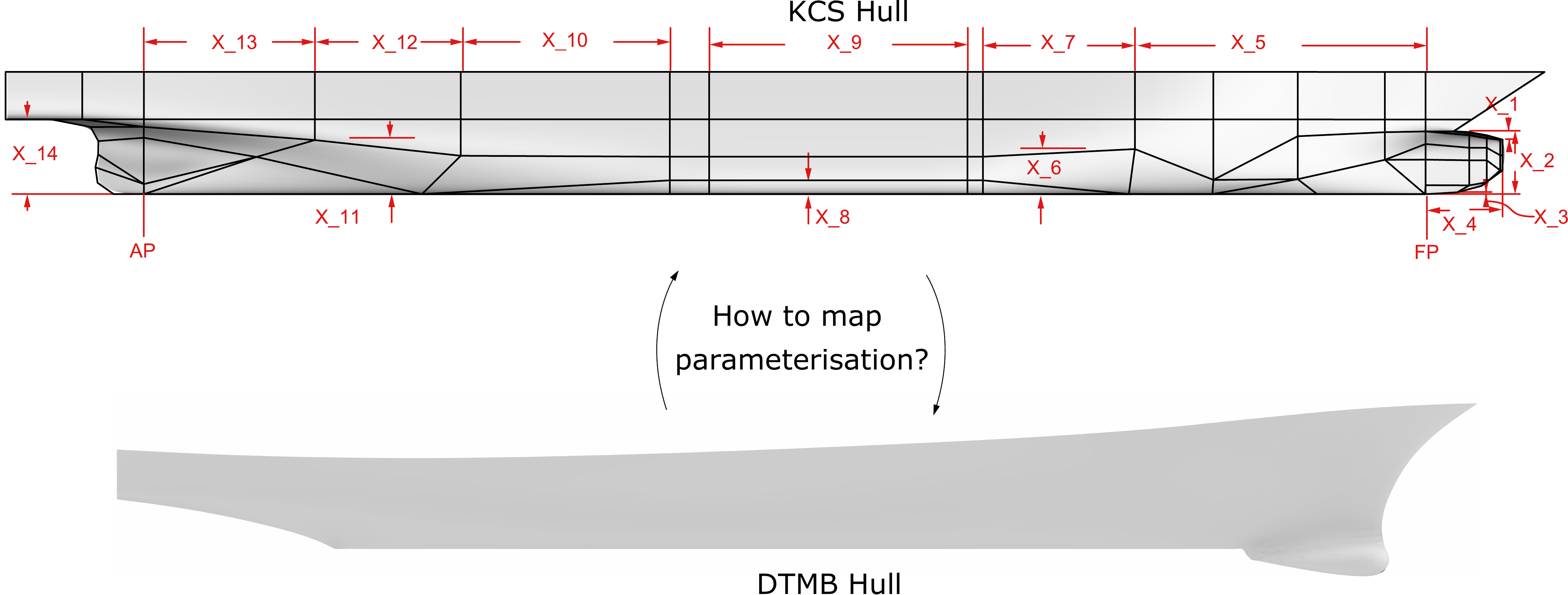}
    \caption{The parameterisation proposed by \cite{intra_r56} for container ship hulls. Is it applicable to a naval ship design such as the DTMB hull?}
    \label{shipgan_f0}
    \end{figure}

    There have been substantial efforts in computer-aided ship design for building robust parametric tools, but they can only handle a specific hull type; some relevant examples of such tools are presented in \cite{GinnisEtAl2011,khan2022modiyacht,intra_r56,khan2019genyacht,khan2017novel}. Despite their efficiency in creating valid and smooth ship-hull geometries, they cannot be readily used to generate instances of ship types that deviate significantly from their target ship types. For example, in Fig. \ref{shipgan_f0}, the parametric construction proposed by \cite{intra_r56}, and later explicitly adapted for container ship hulls by \cite{intra_r17}, is depicted. Such parameterisation cannot be directly or easily mapped to an entirely different ship-hull type, such as the DTMB naval ship shown in the same figure. Although some generic approaches, like FFD (free-form deformation) \cite{intra_r16}, may be applicable to some extent, they either use a rather crude low-fidelity and featureless representation or require significant effort and experimentation for adaptation into new designs. For example, FFD-based parameterisations are not truly feature-driven \cite{gmdsa_r34}, which deprives designers of the commonly needed feature-modelling capabilities and local control for designs such as bulbous bows or other features of local nature. 

\begin{figure}[htb!] 
    \centering
    \includegraphics[width=01\textwidth]{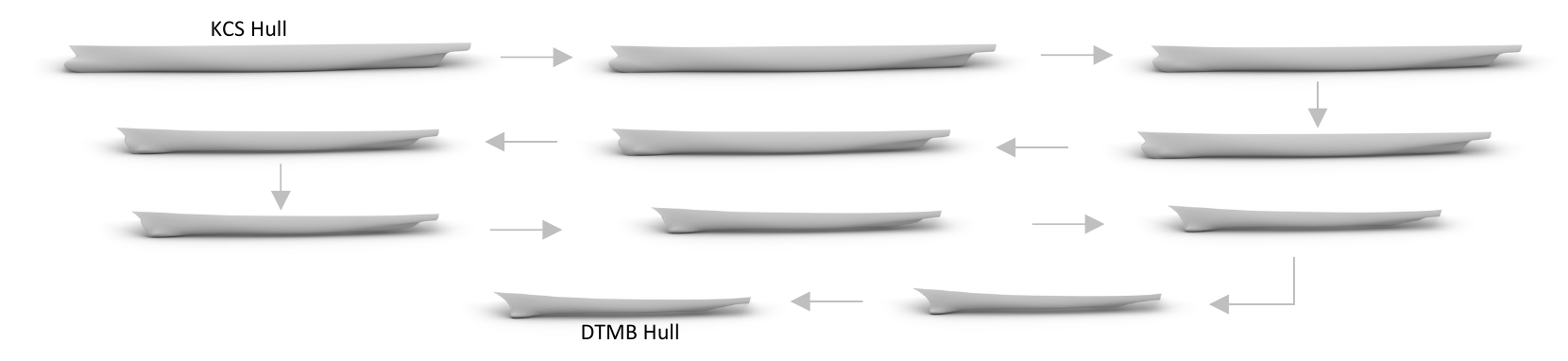}
    \caption{ Transformation of KCS hull into DTMB hull achieved using the ShipHullGAN parametric modeller. Training on synthetic variations of both hulls makes it possible to generate unique designs featuring a blend of KCS and DTMB attributes, exemplified in designs 5-7 along the sequence of arrows.
    }
    \label{shipgan_f1}
    \end{figure}

    In this work, we aim to tackle the above-mentioned challenges in a typical parametric hull design by proposing a generic parametric modeller, ShipHullGAN. The new model can handle various ship hull types and transform one type into a completely different one, as illustrated in Fig. \ref{shipgan_f1}. Additionally, it has the ability to generate unique geometries by augmenting features from different ship types, such as the middle three geometries of Fig. \ref{shipgan_f1}. The proposed modeller is built using deep generative models, specifically deep convolutional generative adversarial networks (GANs) \cite{yu2017unsupervised,li2020efficient}, with a new architecture and loss function suitable for the problem at hand. These generative models were initially proven to be promising for generating entirely novel images from given datasets and recently have been exploited for engineering design problems, i.e., aerodynamic design and optimisation \cite{ssdr_r21,li2020efficient}. If appropriately trained, they can efficiently learn latent representations, which can then be used as design parameters to construct diverse design spaces for shape optimisation. However, the capacity of these approaches has not been explored in ship design. 

    Despite their proven efficiency in design, these models have their limitations. Since they were initially developed for 2D datasets, e.g., processing of images, their application in 3D design requires \textit{suitable geometric representations} to extract meaningful features~\cite{regenwetter2022deep}. An inappropriate training of such models can therefore result in many invalid shapes. More importantly, if the dataset is composed of various design sub-classes, they also tend to lose in generalisability \cite{ssdr_r26}. 
    
    We, therefore, propose a modified architecture and a loss function to overcome the drawbacks inherited from GAN. To commence the training of ShipHullGAN, we first developed a technique to transform different types of ship hulls into a common geometric representation. Furthermore, we constructed a shape-signature tensor (SST) using appropriately encoded designs and their geometric moments (GMs)~\cite{gmdsa_r2}. Therefore, the so-constructed SST augments and enriches the geometric information related to designs given to the ShipHullGAN model during training by infusing the moment-related physics associated with ship hulls. In this way, SST acts as a unique descriptor of each dataset design instance that enables the extraction of meaningful features which are not only geometry-driven but also physics-informed to provide rich and physically-valid design alternatives. We use a deep convolutional architecture \cite{yu2017unsupervised} for the model to capture sparsity in the training dataset, along with a space-filling term~\cite{intra_r46} in the loss function to enhance diversity. 

To the best of the authors' knowledge, this is the first attempt to construct a generic parametric modeller in the field of parametric computer-aided ship design. In accordance with the aim of this work, we report the following main contributions:

\begin{enumerate}
    \item Development of a large shape dataset containing 52,591 \textit{physically validated} design variations of several \textit{existing} classes of ships, some of which are widely used benchmarks in industry and academia. No such extensive dataset of ship hull forms is publicly available.
    \item Development of an intuitive approach to convert all ship designs into a common geometric representation. This technique also ensures a smooth NURBS-based reconstruction of designs resulting from the trained generator. 
    \item The combination of the geometry with its relevant geometric moments results in SST, enabling the capturing of global and local geometric features with physics-informed elements in the latent space, which in turn allows the generation of designs that are both geometrically valid and physically plausible.
    \item Introduction of a space-filling term to the loss function, which enables the model to cover the entire spectrum of the training dataset, thereby enhancing diversity. 
    \item Empirical data from optimisation and comparative studies demonstrating the generic capabilities of ShipHullGAN and its advantage over typical GANs in terms of design diversity, quality, and validity.
\end{enumerate}

\section{Background}
In this section, we provide a concise overview of the current state of parametric ship design, as well as a brief introduction to GANs and their existing applications in engineering design.

\subsection{Parametric ship design}

An early attempt for parametric modelling of ship forms was made by Lackenby \cite{intra_r68}, wherein hull variations were achieved by adjusting the prismatic coefficient, centre of buoyancy, and the dimensions and location of the nearly cylindrical mid-body of a base hull. This method has since evolved by taking on-board tools and representation offered by CAGD (Computer-Aided Geometric Design) to establish the field of Computer-Aided Parametric Ship Design (CAPSD), whose expansion, modernisation and embedding in the area of free-form shape optimisation is arguably due to Horst Nowacki \cite{nowacki2010five} his students and collaborators; see, e.g., \cite{harries1998parametric,kim2004parametric,kaklis2021special} 

In the pertinent literature, various techniques have been employed for the parametrisation of free-form shapes, including direct mesh-, basis vector-, domain element-, conformal mapping, partial differential equation-, FFD-, polynomial and spline-based approaches. A detailed description of these approaches can be found in \cite{samareh2001survey}. However, within the realm of CAPSD, FFD- and spline-based techniques are commonly utilised to parameterise various types of hull geometries.

These methods have been developed to address numerous tasks and overcome specific challenges, such as ensuring hull fairness \cite{pigounakis1996fairing,pigounakis1996convexity}, enhancing design variation \cite{gmdsa_r62}, achieving accurate geometric representation \cite{greshake2018application, GinnisEtAl2011}, enabling plausible hull modifications \cite{perez2011constrained,khan2017novel}, and establishing better associations with solvers \cite{gmdsa_r64, intra_r56}. Despite the diversity of these approaches, the majority of the research has primarily focused on creating parametrisation techniques for specific hull categories and still suffer to overcome problems related to:

\begin{itemize}
    \item Geometric complexity: Ship hulls are characterised by their intricate curvature and non-uniform surfaces, which can make it difficult to develop a comprehensive parameterisation that accurately captures the nuances of the geometry. 
    \item High dimensionality: A ship hull's parameter space can be vast, with numerous parameters governing its form and function. Navigating this high-dimensional space can be computationally demanding, requiring advanced optimisation algorithms and techniques to efficiently explore and evaluate design alternatives.
    \item Interdependencies: The various parameters that define a ship hull's shape and characteristics are often interconnected, with changes in one parameter potentially affecting multiple aspects of the design. 
    \item Constraints: Ship hull designs must adhere to numerous constraints, including physical limitations, regulatory requirements, and industry standards.
    \item Scalability: As marine vessels continue to grow in size and complexity, parameterisation techniques must adapt to accommodate these expanding scales. 
\end{itemize}

Emergence of new geometric approaches and computational intelligence have aided in overcoming a few of the above-mentioned challenges. For example, geometric complexities have been addressed with new surface representations, such as T-splines, which compared to NURBS provide accurate representation and controllability of local and global features \cite{intra_r17}. Issues of high-dimensionality are managed by advanced physics- and geometry-informed approaches \cite{khan2022shape}. Understanding and managing parameter interdependencies is crucial to achieving a successful and coherent design, which is accomplished using procedural approaches that take into account design constraints \cite{khan2022modiyacht,perez2011constrained}.

However, the aspect of scalability remains a significant challenge, which has not been explored extensively within the community. Most of the approaches described above are developed around a baseline from which variations are derived, making it crucial to adapt these methods to accommodate the increasing complexity and demand for innovative and specialised vessels.

In the present work, we aim to develop a generic parametric modeller using GANs (Generative Adversarial Networks) to overcome these limitations. By leveraging the power of GANs, we hope to create a scalable and adaptive solution that can effectively handle the challenges posed by various ship hull designs and cater to the ever-evolving requirements of modern marine vessels.

\subsection{Generative adversarial networks}
This section briefly introduces typical GANs, i.e., Vanilla GAN, and their applications in engineering design and optimisation. A typical GAN model consists of two neural networks, generator $G$ and discriminator $D$, which are trained simultaneously to enhance the capability of $G$ to map from a latent space to the data distribution of interest and thus aim to generate new designs which could have been part of the real designs dataset. In contrast, $D$ tries to classify designs, i.e., to distinguish between real (designs in the training dataset) and generated designs, also referred to as fake designs. Networks $G$ and $D$ are trained simultaneously to reach a Nash equilibrium with the following minimax loss function:

\begin{equation}\label{GANlossFun}
    \min_G\max_D\mathcal{L}_{adv} (D,G) = E_{\mathbf{x}\sim p_{data}(\mathbf{x})}[\log(D(\mathbf{x}))]+E_{\mathbf{z}\sim p_{\mathbf{z}}(\mathbf{z})}[\log(1-D(G(\mathbf{z})))]
\end{equation}

\noindent where $\mathbf{x}$ represents designs in the training dataset and $\mathbf{z}$ denotes the latent tensors randomly sampled from a given distribution $p_{\mathbf{z}}$. The training of the GAN is typically seen as a game or competition between $G$ and $D$, thus referred to as adversarial training, which facilitates learning the data distribution $p_{data}(\mathbf{x})$ of real designs $\mathbf{x}$. During training, the performance of $D$ is maximised so that it can accurately distinguish $\mathbf{x}$ from the synthetic designs, $G(\mathbf{z})$, sampled from $p_{\mathbf{z}}$. During this training, G minimises $\log(1-D(G(\mathbf{z})))$ to learn to produce designs that the discriminator will classify as real designs, i.e., designs resulting from the generator will tend to be similar to real designs.

The adversarial training commences with mini-batches of samples from $p_{\mathbf{z}}$, and $G$ tries to produce realistic designs based on these samples. Then, $D$ is trained to identify whether the presented designs are real (i.e., from the training dataset) or fake (i.e., from the generator). During this process, both networks adjust/optimise their parameters to outperform their opponent, i.e., as $D$ improves its classification ability, $G$ also enhances its ability to create data that fools $D$. This process continues until convergence is achieved. This way, $G$ of the trained GAN model can generate new designs with sufficient diversity within the prior distribution. 

Both $G$ and $D$ can be nonlinear mapping functions, such as a conventional neural network (NN) or a convolutional NN (CNN). In our case, we use CNN, which has been proven more effective in capturing sparse features. $D$ and $G$ with CNN-like architecture are often referred to as Deep Convolutional GAN (DCGAN) \cite{yu2017unsupervised,li2020efficient}.  

\subsubsection{GANs in engineering design}
GANs and their variations have been used for various tasks; however, in this work, we focus on their application in engineering design. Recent applications of GANs in the context of engineering design have appeared in topology optimisation~\cite{oh2019deep,nie2021topologygan}, design and optimisation of aerofoils and wings~\cite{ssdr_r21}, design of metamaterials~\cite{wang2022ih} and synthesis of design creativity in bicycle design~\cite{heyrani2021creativegan}, among many others. 

Chen et al.~\cite{ssdr_r21} proposed a B\'ezier-GAN model for airfoil design and optimisation. To achieve a high representation capacity (i.e., design variation) and compactness (i.e., design validity), B\'ezier-GAN uses a B\'ezier curve layer right after the generator, which fits a B\'ezier curve to data sampled from the employed distribution. Later, Chen et al. in \cite{chen2022inverse} proposed a B\'ezier-GAN variation based on conditional GANs, called CBGAN, to mainly tackle the inversion ambiguity in the inverse design of aerofoils. A performance-conditioned diverse GAN (PcDGAN) was proposed by Nobari et al.~\cite{heyrani2021pcdgan}, which uses a new self-reinforcing score (Lambert Log Exponential Transition Score) for improved conditioning. Chen and Ahmed proposed a performance-augmented diverse generative adversarial network (PaDGAN) \cite{ssdr_r26} and its multiobjective extension MO-PaDGAn \cite{chen2021mo} to ensure that the trained generator remains applicable, with good-performing designs, outside the training dataset domain. To achieve this objective, PaDGAN uses a new loss function based on determinantal point processes (DPPs), which tries to maximise the spread of designs based on their geometric similarity and performance. However, PaDGAN requires the evaluation of performance and its gradients, which is commonly computationally expensive to evaluate. This problem is tackled in the present work using geometric moments (GMs) as a physics-informed performance descriptor instead of directly employing performance evaluations. 

To detect geometric abnormality of generated aerofoils or wings, Li et al., \cite{li2020efficient} trained a DCGAN with a discriminative model based on convolutional neural networks, which detects invalid designs without the need for a separate and expensive computational evaluation. Chen and Fuge \cite{piffl_r5} proposed a hierarchical GAN model to allow the synthesis of designs with interpart dependencies. Nobari et al. \cite{nobari2022range} trained a conditional GAN model to enforce the generator to create designs within a specific performance range and tested their network in the generation of 3D shapes corresponding to aeroplanes. A CreativeGAN model was proposed by Nobari et al. \cite{heyrani2021creativegan} to ensure the generation of novel design alternatives. To enhance novelty, CreativeGAN used the K-nearest neighbour (KNN) approach to detect novel features of designs and use these features to train the StyleGAN model \cite{karras2019style}, which is capable of generating designs with the detected novel features. Lastly, Dong et al. \cite{dong2023shipgan} demonstrated a non-design application of GANs by developing ShipGAN, a model that generates realistic operational scenarios for ships.

\section{ShipHullGAN}\label{sec:ShipHullGAN}
\begin{figure}[htb!]
    \centering
    \includegraphics[width=01\textwidth]{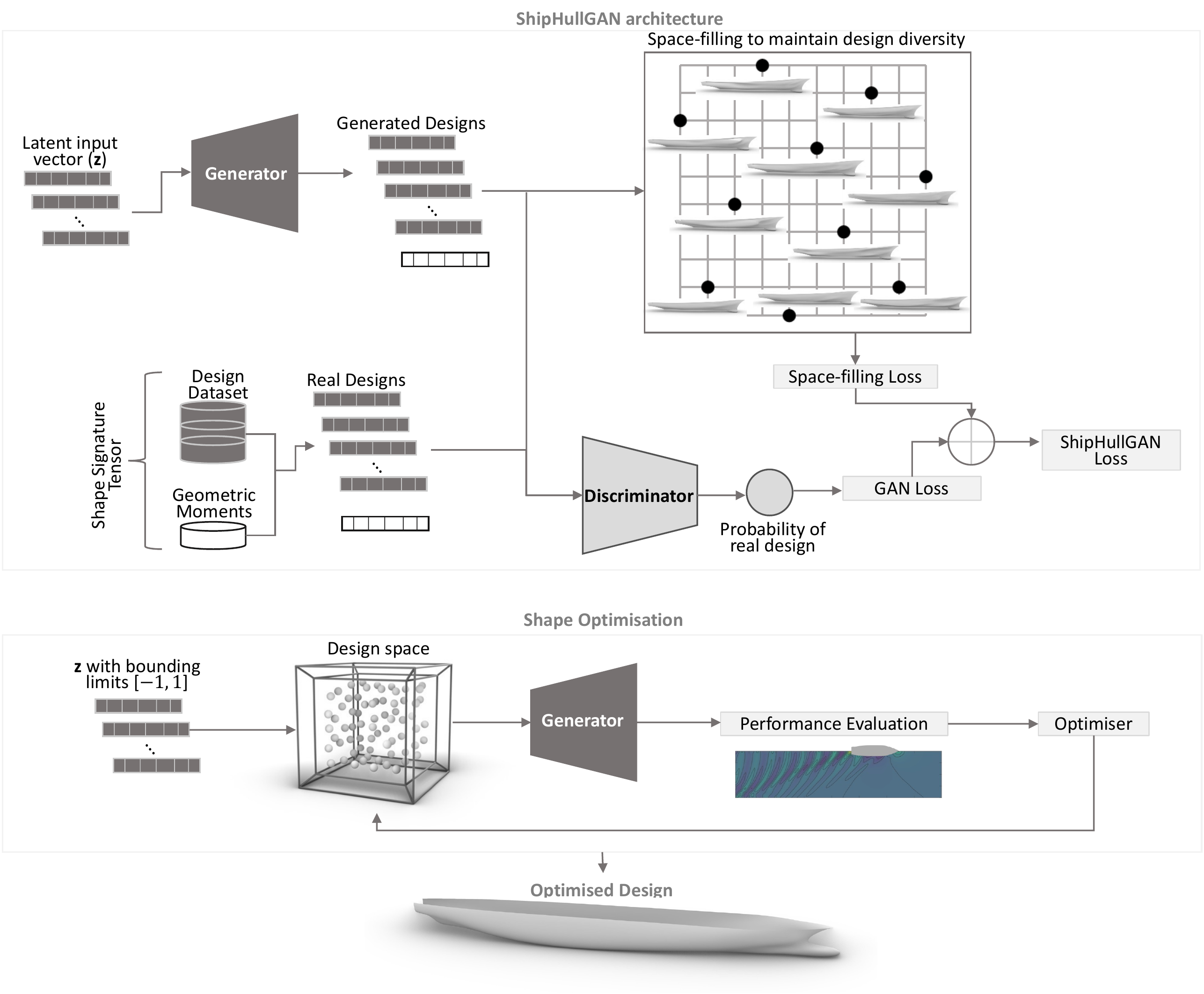}
    \caption{ The ShipHullGAN architecture incorporates shapes and their geometric moments in the form of SST to improve design validity and incorporate physics into the latent variables. It also includes a space-filling layer that aims to create a uniform distribution of designs from the generator. Once trained, the generator can then be linked with the performance evaluation code and optimiser to perform shape optimisation for optimised design alternatives satisfying given design constraints.}
    \label{shipgan_f7}
    \end{figure}

In this section, we provide an in-depth presentation of the ShipHullGAN model considerations and its architecture, schematically depicted in Fig.~\ref{shipgan_f7}. The generator and discriminator of the proposed model have a deep convolutional architecture to better capture the sparsity in the data. ShipHullGAN uses space-filling~\cite{intra_r46} to evenly capture the diversity present within the training dataset and SST to inject the notion of physics in the latent features during training. 

Let ${\cal G}$ be a geometric object representing a baseline design (e.g., a parent hull) in an ambient space ${\cal A}\subseteq\mathbb{R}^3$.
We also assume that $\bm{P}({\cal G})$ is a  vector function in a finite space that provides the GAN suitable geometric representation of ${\cal G}$, $\mathbf{x} = \bm{P}({\cal G})$, in $\mathcal{A}$. Along with $\mathbf{x}$, there is a lumped geometric moment vector, $\bm{M}({\cal G}) \in \mathbb{R}^{n_M}$. Now combining the geometry and its moments results in a unique SST,
    \begin{equation}\label{ssdr_e3}
       \mathrm{SST} =\left(\bm{P}({\cal G}),\bm{M}({\cal G})\right),
    \end{equation}

\noindent encompassing high-level information about the design. If the shape dataset contains $\{$\mathlist{\mathbf{x}_1,\mathbf{x}_2,\mathbf{x}_3,\dots,\mathbf{x}_n}$\}$ designs, then computing the GMs of each design results in a training dataset with $n$ SSTs, denoted as $\mathcal{X}=\{$\mathlist{\mathrm{SST}_1,\mathrm{SST}_2,\mathrm{SST}_3,\dots,\mathrm{SST}_n}$\}$, for training the ShipHullGAN model.

\subsection{Shape dataset}
\begin{figure}[htb!]
    \centering
    \includegraphics[width=01\textwidth]{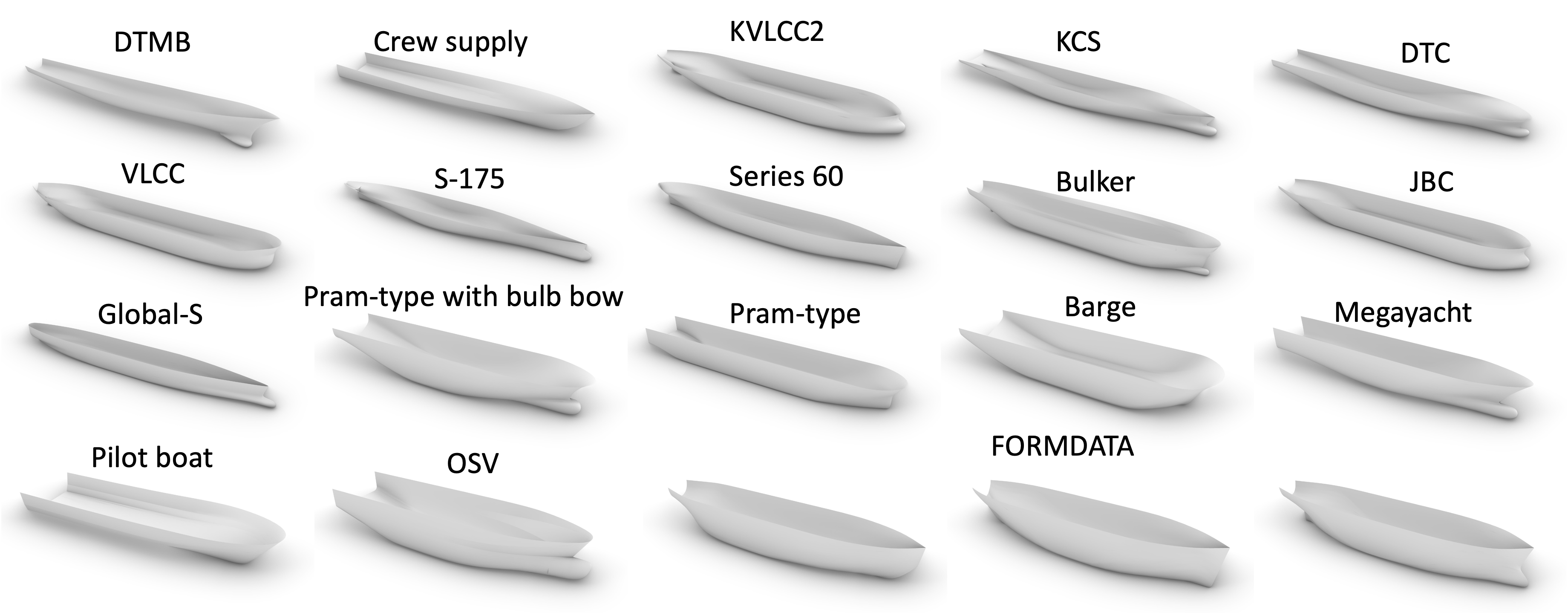}
    \caption{Main ship hull types used in training of ShipHullGAN model. }
    \label{shipgan_f2}
    \end{figure}

SciML for engineering design problems suffers mostly from inappropriate and/or insufficient amounts of data. This is especially challenging if labels, typically performance parameters, are evaluated by time-consuming high-fidelity solvers. However, generative models are generally unsupervised and do not require labels; nevertheless, a sufficiently diverse dataset with novel design alternatives is necessary to acquire a trained model with good generalisability. In the context of engineering design, application of such models has so far appeared in automotive \cite{radhakrishnan2018creative} and aerofoil \cite{ssdr_r21} design, since relevant datasets such as shapeNet\footnote{\url{https://shapenet.org}} and UIUC airfoil coordinates database\footnote{\url{https://m-selig.ae.illinois.edu/ads/coord_database.html}}, containing several thousand designs, are publicly available. To the best of the authors' knowledge, no equivalent, diverse and publicly available dataset of ship-hull designs exists. This is probably why so far in ship design research, SciML models are implemented on a specific design type whose variations are created synthetically using a baseline parameterisation. However, in such cases, new hulls are generally slight variations of the parent hull (baseline design). Therefore, if GANs were trained on a specific ship-hull type using a similar baseline variation process, one could not expect significant novelties in generated designs. To overcome this hurdle and construct a sufficiently diverse and large dataset of existing ship-hull geometries, we extensively studied the pertinent literature on systematic hull form series, optimisation, and machine learning to extract all relevant hull types. This exercise resulted in consideration of systematic series, e.g. FORMDATA, and a variety of parent hull families from different ship types, e.g., KCS\footnote{\url{http://www.simman2008.dk/KCS/kcs_geometry.htm}}, KVLCC2\footnote{\url{http://www.simman2008.dk/kvlcc/kvlcc2/kvlcc2_geometry.html}}, VLCC, JBC\footnote{\url{https://www.t2015.nmri.go.jp/jbc.html}}, DTC, and DTMB\footnote{\url{http://www.simman2008.dk/5415/combatant.html}}), shown in Fig. \ref{shipgan_f2}, which are widely used in industry and academia.  

\begin{figure}[htb!]
    \centering
    \includegraphics[width=01\textwidth]{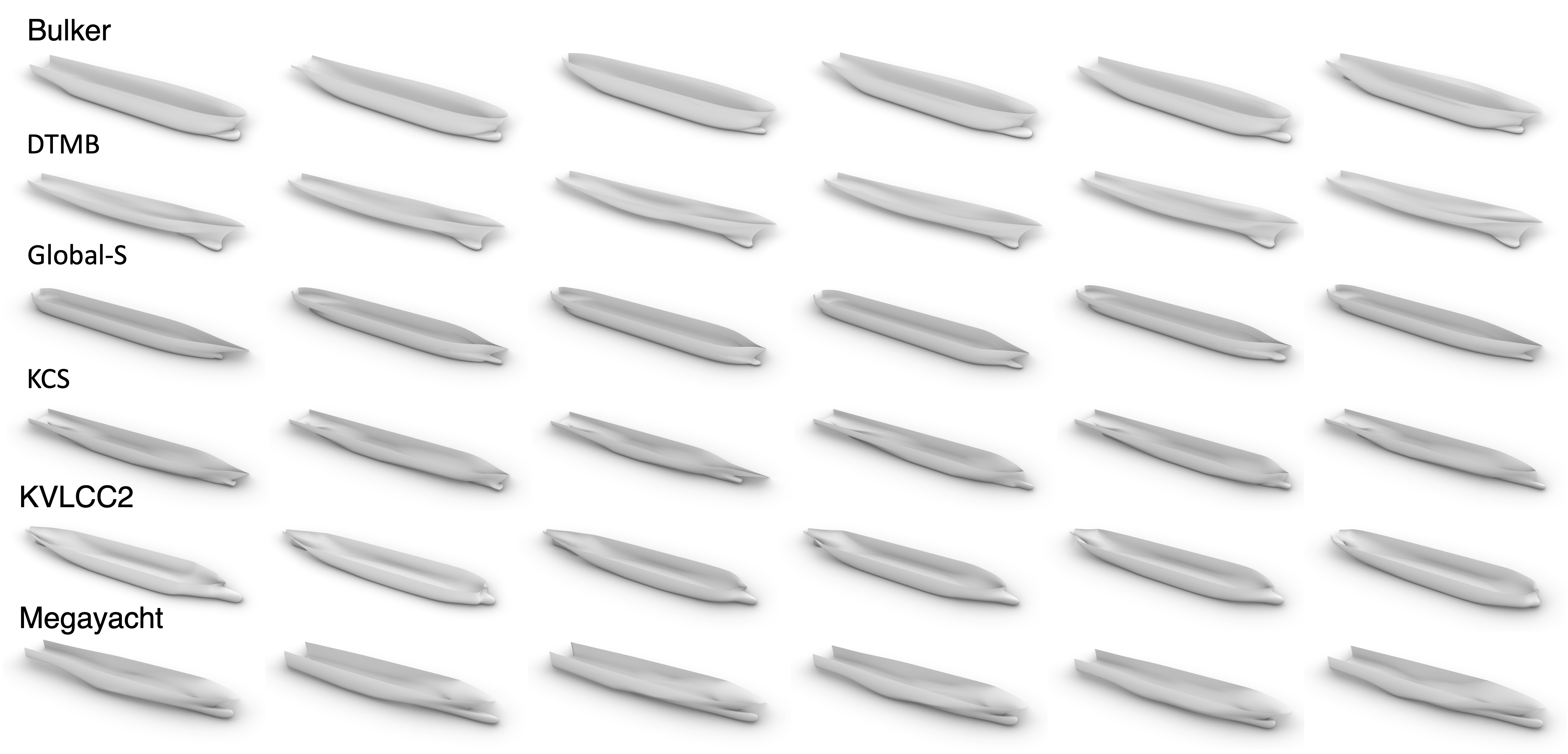}
    \caption{Indicative instances from the synthetic design variation of \textit{Bulker, DTMB, Global-S, KCS, KVLCC2,} and \textit{Megayacht} hulls in Fig.~\ref{shipgan_f2} created for training ShipHullGAN.}
    \label{shipgan_f3}
    \end{figure}

\begin{figure}[htb!]
    \centering
    \includegraphics[width=01\textwidth]{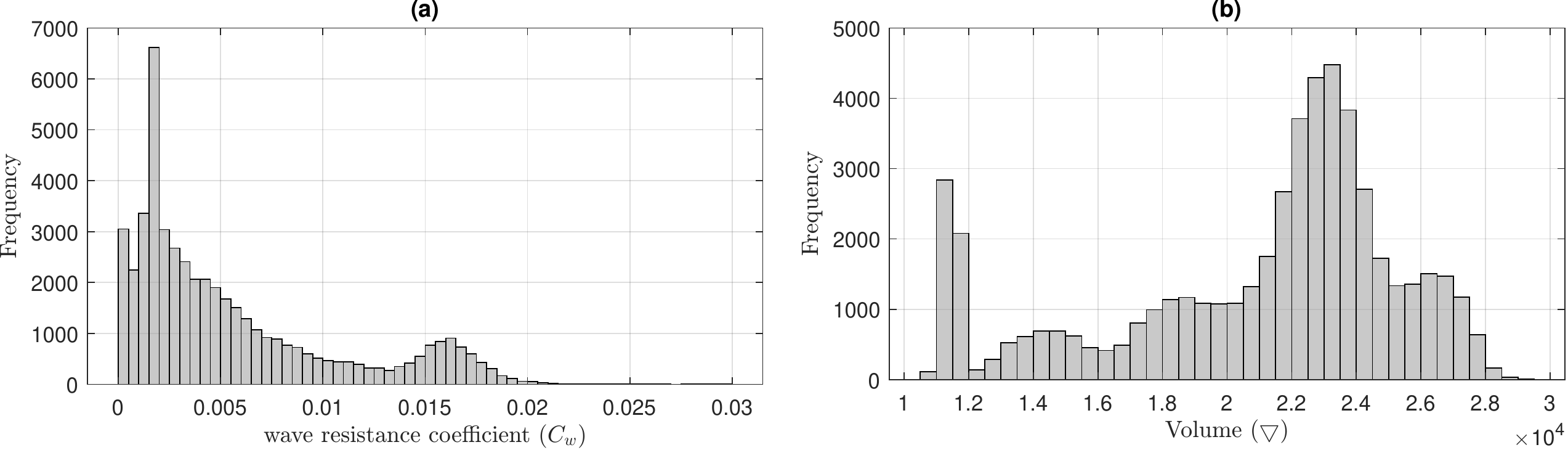}
    \caption{Distribution of (a) wave resistance coefficient ($C_w$) and (b) ship-hull volume ($\bigtriangledown$) in the training dataset.}
    \label{shipgan_f9}
    \end{figure}

Among the hulls in Fig. \ref{shipgan_f2}, the FORMDATA series~\cite{FORMDATA63} is based on a systematic analysis of geometric data of a high number of existing ships in the 1960's and of earlier systematic series, covers conventional, mainly wall-sided hull forms, and has been widely used for designing merchant ships. Hull variations from the FORMDATA series can supply us with approximately 5000 different hull forms, but of only three basic ship hull types, referred to as U, N and V, which are generated by combining different groups of ship sections for the aft and fore parts. The shapes of these ship lines are varied systematically using three form coefficients, i.e., midship section coefficient $C_M$, along with the fore $C_{B_F}$, and aft $C_{B_A}$ block coefficients. Therefore, the FORMDATA-generated dataset requires additional designs to meet the diversity requirements discussed before. If we merely add the previously identified additional parent hull geometries, we expect no or minimal impact as they will constitute a negligible percentage of the dataset. To overcome this issue, we created synthetic variations of the remaining hulls in Fig. \ref{shipgan_f2} based on the parametric approach discussed in \cite{gmdsa_r34}. These designs' length, beam and width are kept constant, and non-dimensional shape parameters, varying between 0 and 1 (0 to 100\%), are used to create valid ship-hull shape variations. Indicative instances of the variations accomplished by this approach are shown in Fig.~\ref{shipgan_f3}. It can be easily seen that all depicted hull instances have plausible geometries with non-negligible variation when compared to the parent hull design. Aggregating the full set of parent-hull design variations with the FORMDATA-generated designs result in 52,591 designs which are then used to train the ShipHullGAN model. Finally, the distribution training designs' physical (i.e., wave resistance ($C_w$)) and geometric (i.e., volume ($\bigtriangledown$)) criteria are shown in Fig.~\ref{shipgan_f9}. The distribution of physical (i.e. wave resistance $C_w$) and geometric (i.e. volume $\bigtriangledown$) criteria of the training designs are illustrated in Fig.~\ref{shipgan_f9}. The distribution of $\bigtriangledown$ indicates that our dataset has adequate diversity, with most designs having minimal $C_w$. However, these distributions do not play any direct role in the output of ShipHullGAN, as training is performed with only design geometries. 

\subsection{Shape encoding for GANs}
\begin{figure}[htb!]
    \centering
    \includegraphics[width=01\textwidth]{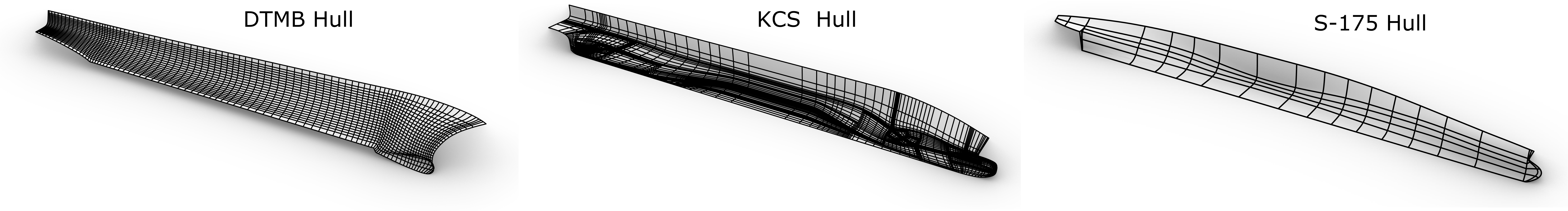}
    \caption{Example of three ship hulls with different structures of the surface parameterisation: the DTMB hull is constructed with a single NURBS surface, whereas the KCS and S-175 are composed of several NURBS surface patches with a significantly different number of control points.}
    \label{shipgan_f4}
    \end{figure}

Typically, deep learning models require datasets with vector inputs of fixed dimensions to extract meaningful features. This is relatively easy to achieve for natural language processing and vision/image processing, where these models originated. However, selecting suitable data encoding is a significant challenge when considering applications of deep learning models in 3D free-form shape processing. Free-form shapes, even when belonging to the same family, can have significantly different topology, structure, geometric parameterisation, and resolution; see Fig.~\ref{shipgan_f4} as an example of three ship hulls with significantly different geometrical representations and surface dimensionality. Therefore, we need to ensure that all shapes in the training dataset share the same underlying topology, representation, and resolution. This implies that all designs need to be converted into a common representation with a similar resolution at a prepossessing stage. 

Signed distance function (SDF), voxels, point clouds and meshes are commonly used with satisfactory results for shape visualisation tasks in computer graphics and machine learning-based regression models for performance prediction~\cite{regenwetter2022deep}. 
In generative models, however, where the output is also a 3D shape, these approaches often result in the loss of local geometric features of the input shapes. More importantly, the resulting designs of such approaches commonly lack surface smoothness, which is crucial for several engineering analyses. In the case of ship hulls, both local features and surface smoothness are essential in appropriately evaluating the hydrodynamic performance of a ship hull. Although one can achieve a certain level of smoothness by increasing the employed resolution, this also increases the network complexity and memory requirements. A detailed discussion of such approaches with their advantages and disadvantages can be found in~\cite{regenwetter2022deep}.

NURBS-based surface representations are quite common among ship hull designers as they provide the most accurate and versatile mathematical description of design geometry and are thus favoured in the pertinent industry and the vast majority of CAD tools available. As mentioned before, DCGAN models require fixed dimensional vectors as input, and therefore a common description is needed. However, especially when the dataset comprises different design classes, converting all of them into a common NURBS representation is not a trivial task, especially for 3D shapes.

In summary, any approach used for the construction of a 3D dataset for SciML training should:

\begin{enumerate}
    \item Represent all shapes with the same resolution;
    \item Capture both local and global geometric features of the shape;
    \item Maintain geometric similarity between the original and reconstructed shapes;
    \item Satisfy the above conditions with a relatively low resolution (e.g., with few mesh elements) to avoid redundancies and reduce the model's overall complexity.
\end{enumerate}

\begin{figure}[htb!]
    \centering
    \includegraphics[width=01\textwidth]{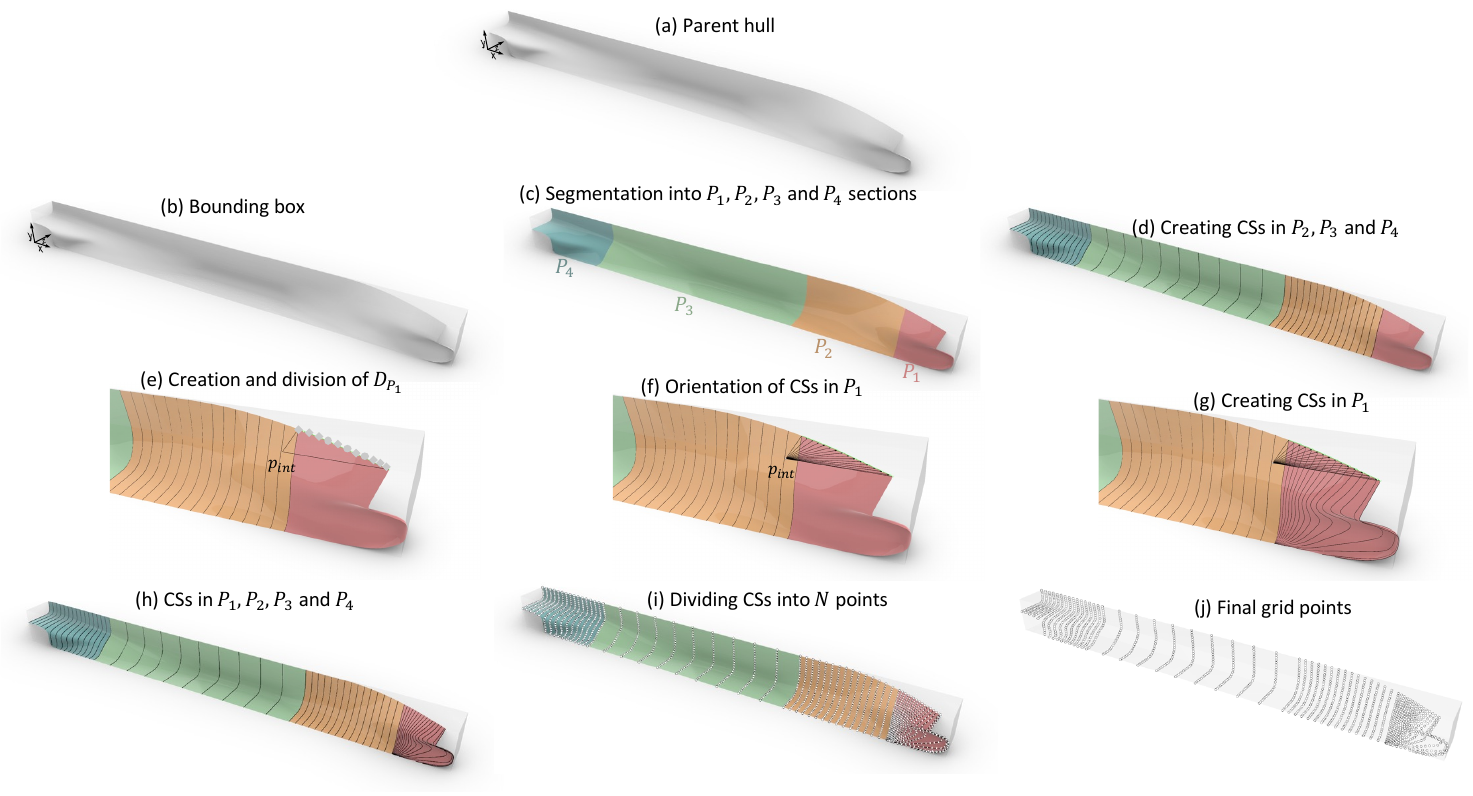}
    \caption{Steps of the proposed body-plan-based approach for extracting geometric information from ship-hull shapes.}
    \label{shipgan_f5}
    \end{figure}

 In traditional and even modern ship design, the Body Plan (BP), consisting of the so-called cross sections (CSs) resulting from the intersection of the ship hull with an appropriate sequence of transverse planes along the length of the ship, is a handy representation of ship's geometry in $2.5D$ format. If appropriately constructed, it can be used, along with the basic reference lines, to develop the remaining ship lines plans, i.e.,  the profile plan, the waterlines (intersection of ship hull with horizontal planes) and the buttocks (intersection with planes parallel to the symmetry plane) of the ship. Therefore, a BP-inspired approach can encode the geometric information in a uniform and consistent manner in a ship hull. Our approach is based on the intuitive arrangement of transverse planes used for producing the BP so that all critical features of the hull surface are captured. More importantly, once a new design is generated from the GAN model, we can reconstruct a smooth and fair hull surface with sufficient accuracy and relative ease. The basic steps of our implementation are summarised below and illustrated in Fig.~\ref{shipgan_f5} for the KCS hull case. 

\begin{enumerate}
    \item Assume that the ship hull is placed within the smallest axis-aligned bounding box as shown in Fig.~\ref{shipgan_f5}(b) with $\bar{L}, \bar{B},$ and $\bar{D}$ denoting its longitudinal, transverse and vertical dimensions, respectively. 
    \item Using the bounding-box length $\bar{L}$, convert the hull geometry into a non-dimensional representation contained in a bounding box with dimensions $1, \frac{\bar B}{\bar L},$ and $\frac{\bar D}{\bar L}$. 
    \item Divide the hull into four parts using a non-uniform partition, $[0,0.1,0.3,0.8,1]$, which corresponds to the typical regions of different geometric variation for ship hulls in the longitudinal direction; see Fig.~\ref{shipgan_f5}(c). The intervals $P_1=[0,0.1]$, $P_2=[0.1,0.3]$, $P_3=[0.3,0.8]$, and $P_4=[0.8,1]$ correspond to the bow, fore transition, wall-sided (midship), and stern parts, respectively.
    \item Assuming that $E = 4\bar{E}$ is the overall number of ship CSs used to describe each ship hull in the dataset, where $\bar{E}\in\mathbb{Z}_{>1}$. We divide each region, $P_1$, $P_2$ $P_3$ and $P_4$ into $\frac{E}{4}$ equally spaced CSs. This arrangement generates a dense line description in areas with abrupt geometrical changes ($P_1$, $P_2$, and $P_4$) and a rather sparse representation for the region ($P_3$)with an almost constant CS; see Fig.~\ref{shipgan_f5}(d).
\end{enumerate}

    \noindent The CSs used in our encoding for $P_2$, $P_3$, and $P_4$ correspond to intersections of the ship hull surface with transverse planes, i.e., planes perpendicular to the longitudinal direction, which is the standard practice in ship design. However, CSs in $P_1$ are generated by a family of planes rotating gradually through a vertical axis lying on the intersection of the longitudinal symmetry plane and the transverse plane at $\bar{L}=0.1$, as shown in Figs.~\ref{shipgan_f5}(h-j). This approach is adopted in order to avoid multiply (usually doubly) connected CSs resulting from intersections of the bulbous bow area with transverse planes. In more detail, the following steps describe the construction of CSs in $P_1$: 
\begin{enumerate}
    \item Create the deck curve $D_{P_1}$ of the hull part in $P_1$ and divide it into $\frac{E}{4}$ equally-spaced points using the arc length method; see Fig.~\ref{shipgan_f5}(e)).  
    \item Find intersection point, $p_{int}$, of lines starting from the first and last points of $D_{P_1}$, respectively, along the longitudinal plane of symmetry and the transverse plane at $\bar{L}=0.1$; see Fig.~\ref{shipgan_f5}(e).
    \item Using the line segments defined by $p_{int}$ and each of the identified points on $D_{P_1}$, generate $\frac{E}{4}$ planes intersecting the ship hull; see Fig.~\ref{shipgan_f5}(f).
    \item Create CSs in $P_1$ by computing the intersections of the previously constructed planes and the ship hull; see Fig.~ \ref{shipgan_f5}(g).
\end{enumerate}

\noindent Once CSs in all the regions are created, we divide each CS into $N$ equally-spaced points using the arc length method, see Figs.~\ref{shipgan_f5}(i, j), which results in overall $N\times E$ grid points for each design in the training dataset. 

\begin{figure}[htb!]
    \centering
    \includegraphics[width=01\textwidth]{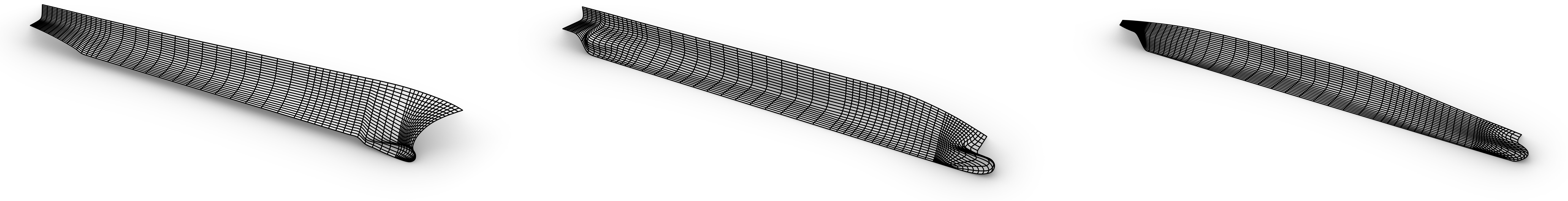}
    \caption{Reconstruction of designs in Fig. \ref{shipgan_f4} using the proposed shape encoding approach. It is evident that all three designs now possess a consistent underlying geometric representation.}
    \label{shipgan_f26}
    \end{figure}

The grid points of a design resulting from the trained generator are used to reconstruct the surface by fitting a curve on the points for each CS, followed by interpolating the surface on the curves, as depicted in Fig. \ref{shipgan_f26}. This surface reconstruction process from grid points is further discussed in detail in \S \ref{designRecinsSec}.
    
\subsection{Preparing geometric data for training}
\begin{figure}[htb!]
    \centering
    \includegraphics[width=01\textwidth]{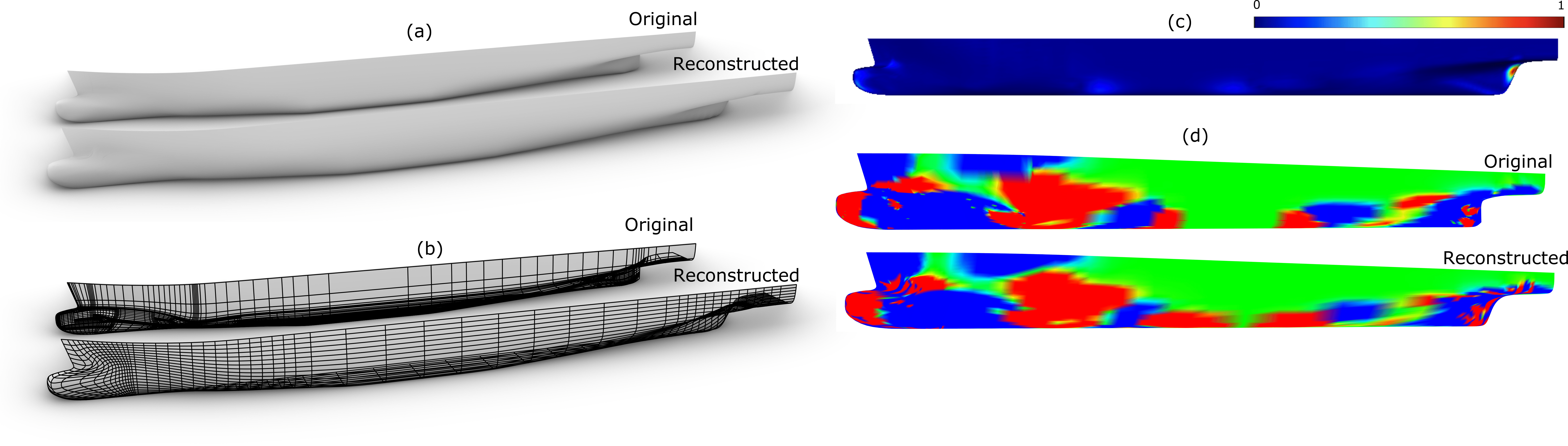}
    \caption{Comparison between the original KCS hull and its surface reconstruction from the grid points of the proposed body-plan-based approach. (a) Surface representations of the original and reconstructed hulls, (b) their geometric representation,  comparisons in terms of (c) the one-sided Hausdorff distance~\cite{cignoni1998metro}, and (d) Gaussian curvature.}
    \label{shipgan_f6}
    \end{figure}

As previously mentioned, there are $n=52,591$ designs in our shape dataset. Before training, all designs in this dataset are deconstructed using the previously described body-plane-based approach. For this deconstruction, we use $E=56$ CSs, and each CS is divided into $N=25$ points. Hence, the $i^{th}$ design will be represented with $\mathbf{x}_i$, corresponding to $25\times 56$ grid points. We have experimented with different grid resolutions, but, as indicated in Fig.~\ref{shipgan_f6}, the employed, relatively low, resolution of $25\times 56$ grid points provide sufficient surface reconstruction accuracy while preserving both local and global geometric features. 

\begin{figure}[htb!]
    \centering
    \includegraphics[width=0.7\textwidth]{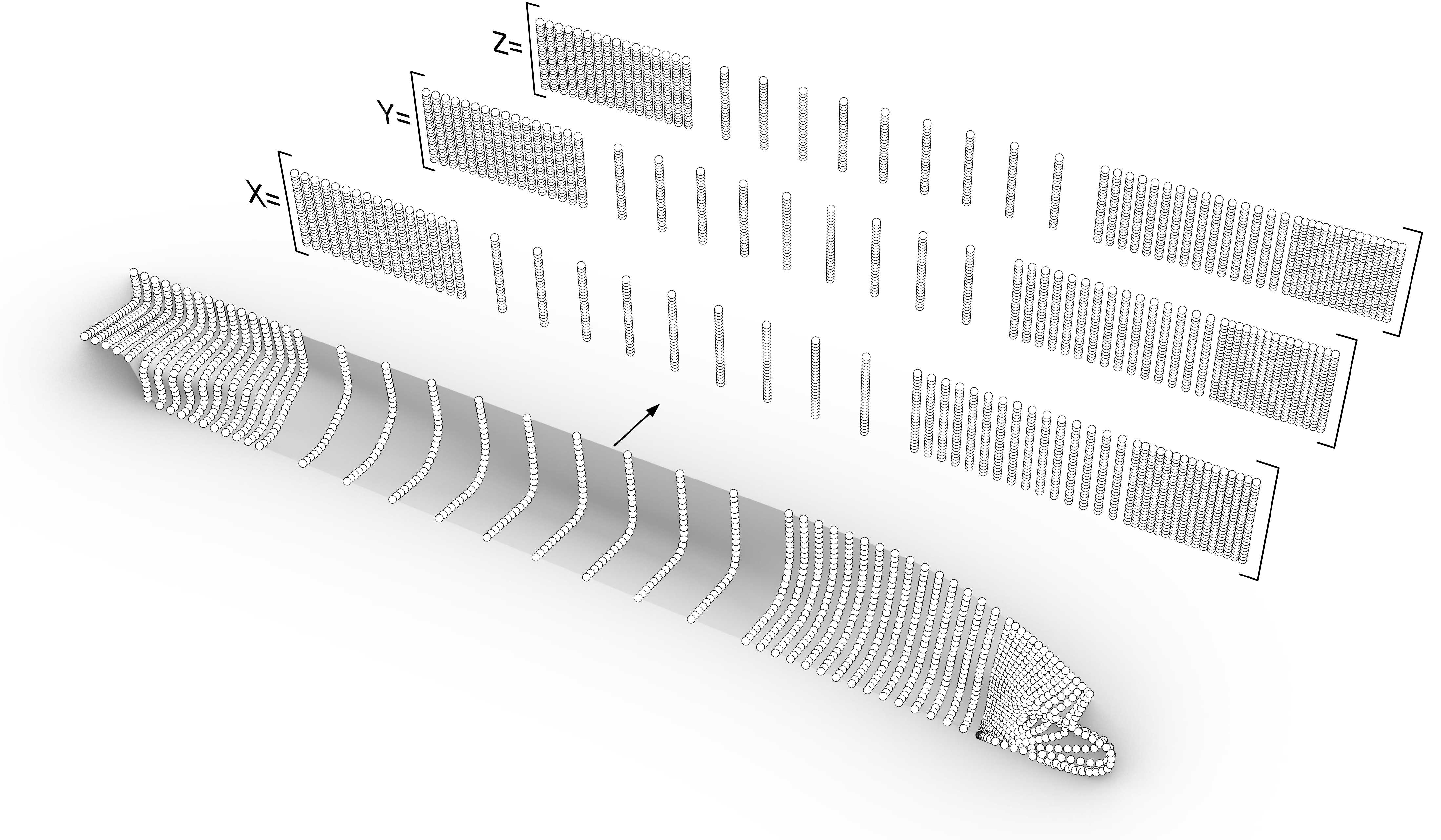}
    \caption{Illustration of transformation of grid points into training set's 3-tuples of input matrices. }
    \label{shipgan_f12}
    \end{figure}

Finally, the $x$ (longitudinal), $y$ (transverse) and $z$ (vertical) coordinates of the generated grid points are used to construct three  $[25\times 56]$ matrices as shown in Fig.~\ref{shipgan_f12}. Hence, the geometric representation/encoding of the shape dataset is materialised with $n=52,591$ 3-tuples of $[25\times 56]$ matrices. The proposed shape encoding approach provides a robust representation of training designs, enabling our model to efficiently learn complex relationships between the input distribution and the generated designs.

\subsection{Enhancing model robustness}
The augmentation of the geometric information in the SST with geometric moments (GMs) significantly increases the model robustness since it enriches the design encoding and subsequently results in fewer invalid and similar shapes. More importantly, due to the existence of a strong correlation between physical quantities of interest (QoI), such as $C_w$ and GMs, the inclusion of the latter in SST constitutes an indirect introduction of physics-related information in the extracted latent features; further details on this procedure will be presented later in the section.

\subsubsection{Geometric moments - GMs}\label{s_qm} 
Assuming an object $\mathcal{G}$, the $s=p+q+r$ order GMs of its shape can be calculated as

	\begin{equation}\label{gmdsa_e2}
		M^{p,q,r}(\mathcal{G}) = \int_{-\infty}^{+\infty}\int_{-\infty}^{+\infty}\int_{-\infty}^{+\infty} x^p~y^q~z^r~\rho(x,y,z)~\text{d}x\text{d}y\text{d}z, \qquad \mathrm{with} \qquad p,q,r\in\lbrace 0,1,2,\dots\rbrace,
	\end{equation}
	
 \noindent where $\rho(x,y,z)=\left\{\begin{array}{cc}1 & \mathrm{if}\,(x,y,z)\in\mathcal{G}\\ 0 &\mathrm{if}\,(x,y,z)\notin\mathcal{G}\end{array}\right.$. Given a non-negative integer $s$, the vector ${\bf M}^s$ will contain $n_M=(s + 1)(s + 2)/2$ moment elements. Ideally, the selection of $s$ should result in a set of GMs capturing global as well as local features of $\mathcal{G}$. For instance, GMs of order 2  are
	
	\begin{equation}\label{gmdsa_e3}
	    \bm{M}^2 = 
	    	\begin{Bmatrix}
	   	M^{2,0,0}(\mathcal{G})&M^{0,2,0}(\mathcal{G})&M^{0,0,2}(\mathcal{G})&M^{1,1,0}(\mathcal{G})&M^{1,0,1}(\mathcal{G})&M^{0,1,1}(\mathcal{G})
	   \end{Bmatrix}.
	\end{equation}	
	
	In Eq.~\eqref{gmdsa_e2}, if $\rho(x,y,z)$ corresponds to volume or mass density, then the zero- and first-order moments, $M^{0,0,0}(\mathcal{G})$, $M^{1,0,0}(\mathcal{G})$, $M^{0,1,0}(\mathcal{G})$, and $M^{0,0,1}(\mathcal{G})$, correspond to commonly used moments in computer graphics, CAD and engineering for computing the object volume or mass, $\mathcal{V}=M^{0,0,0}(\mathcal{G}),$ and the coordinates of the volume or mass centroid. If $\rho(x,y,z)$ is the probability density function of a continuous random variable, then $\bm{M}^0$, $\bm{M}^1$, $\bm{M}^2$, $\bm{M}^3$ and $\bm{M}^4$, represent the total density, mean, variance, skewness and kurtosis of the random variable, respectively. Moreover, the 2nd order GMs can be organised in a second-rank tensor, the moment of inertia tensor.     
    As one might expect, the more GMs we use, the better we capture the shape's intrinsic features. Therefore, one may opt for the inclusion of up to $s$ order moments, i.e., $\left\{\bm{M}^0, \bm{M}^1, \bm{M}^2, \dots, \bm{M}^s\right\}$, with $s$ being appropriately large to cover the shapes of interest~\cite{gmdsa_r8}. Theoretically, $s$ ranges from $0$ to $\infty$, though there exist object classes for which $s$ is finite when, e.g., dealing with the class of the so-called quadrature domains in the complex plane \cite{gmdsa_r58} or when approximating convex bodies using Legendre moments \cite{gmdsa_r59}.
    
    There exists a variety of methods available in the literature for computing GMs, which use either lower-order approximating meshes\cite{gmdsa_r24} or high-order surface representations~\cite{gmdsa_r5} of $\mathcal{G}$, such as B-splines and NURBS. The most commonly used method employs Gauss's divergence theorem, also used in this work. The divergence theorem evaluates GMs by converting volume integrals to integrals over the surface bounding the volume; for further details, the interested reader may refer to \cite{gmdsa_r25}.

    \subsubsection{Geometric moment invariants - GMIs} 
    The GMs discussed so far are not invariant to affine transformations, such as translations, rotations and scaling \cite{gmdsa_r2}. However, most physical quantities are invariant to all or some of these transformations, and we need to match this invariance when using moments to establish the relationship with the corresponding physical quantity. For instance, evaluating $C_w$ for the ship is invariant to translation and scaling if assessed at a certain Froude number. Therefore, we need to ensure the translation/scaling invariance of the moments we will employ in the construction of the SST. This is accomplished with appropriate geometric moment invariants that are briefly discussed subsequently; a more general discussion of GMIs can be found in~\cite{gmdsa_r2}.
   
    If we ensure that the computation presented in Eq.~\eqref{gmdsa_e2} is performed with respect to an origin placed at $\mathcal{G} $'s centroid, $\mathbf{c}(\mathcal{G})=(C_x,C_y,C_z)$, we then get the so-called \textit{central GM} of $s$th order, which is invariant to translation and is equivalently computed as: 
        
	\begin{equation}\label{gmdsa_e6}
		\mu^{p,q,r}(\mathcal{G}) = \int_{-\infty}^{+\infty}\int_{-\infty}^{+\infty}\int_{-\infty}^{+\infty} (x-C_x)^p~(y-C_y)^q~(z-C_z)^r~\rho(x,y,z)~\text{d}x\text{d}y\text{d}z.
	\end{equation}

    \noindent It is worth noting that as this computation is performed with respect to the object's centroid, the first-order moment is zero, i.e., $\left\{\mu^{1,0,0},\mu^{0,1,0},\mu^{0,0,1}\right\}=0$. To further achieve invariance of $\mu^{p,q,r}$ to scaling, we assume that $\mathcal{G}$ is uniformly scaled by a factor $\lambda$, which yields 

	\begin{equation}\label{gmdsa_e7}
    \hat{\mu}^{p,q,r}(\hat{\mathcal{G}}) = \lambda^{p+q+r+3}\mu^{p,q,r}(\mathcal{G}).
	\end{equation}

    \noindent Then, one can easily conclude that  
    \begin{equation}\label{gmdsa_e8}
    MI^{p,q,r}=\frac{\mu^{p,q,r}}{{(\mu^{0,0,0})}^{1+(p+q+r)/3}}
    \end{equation}

\noindent is an invariant moment form of $\mathcal{G}$ under uniform scaling and translation \cite{gmdsa_r2}. For any non-negative integer, $s$, the GMI vector, $\bm{MI}^s$, contains all the geometric moments invariant to translation and scaling such that $p+q+r=s$. By definition this invariance satisfies the following equalities: $MI^{0,0,0}=1$ and $\bm{MI}^1=\{$\mathlist{MI^{1,0,0}, MI^{0,1,0}, MI^{0,0,1}, MI^{1,1,0}, MI^{1,0,1}, MI^{0,1,1}}$\}=\mathbf{0}$.

\subsubsection{Relationship of geometric moments to physics}\label{CwWithMI}
Our motivation to investigate the utility of GMs for SST stems from the extensive use of the \textit{Sectional Area Curve - SAC} and its moments in Computer-Aided Ship Design for hydrostatic and hydrodynamic analyses. SAC is a function $S(x)$ of 2D zeroth-order GMs describing the longitudinal variation of the area of ship sections below the waterline. As stated in~\cite{gmdsa_r61}, \textit{\say{A SAC provides a practical and straightforward description of global geometric properties. At the same time, it is closely related to a ship's resistance and propulsion performance. From this point of view, the ship hull form distortion approach based on SAC transformation is one of the most influential global design methods for the preliminary design stage.}} In an analogous spirit, \cite{gmdsa_r62} stresses that \textit{\say{geometric properties of SAC have a decisive effect on the global hydrodynamic properties of ships}}. Historically, the importance of SAC in ship design was established back in the 1950s with the introduction of the Lackenby transformation~\cite{intra_r68} for modifying the ship hull via the SAC, which has been further enriched in the context of modern CAD representations and used in ship-design optimisation; see, e.g., \cite{gmdsa_r64,gmdsa_r65}.
    
Furthermore, linear wave-resistance analysis performed by eminent hydrodynamicists, such as E.O. Tuck \cite{gmdsa_r66,gmdsa_r67}, J.V. Wehausen \cite{gmdsa_r68} and others, has revealed the importance of the longitudinal rate of change of the cross-sectional area, i.e., $S' (x)$, which determines the strength of the Kelvin-source distribution used to model the disturbance caused by the body as it moves on the sea's free-surface. It is worth noticing that the flow around a slender ship cruising on the free surface with a constant velocity can be modelled by an appropriate source-sink distribution along its centre plane. The strength of these sources is proportional to the longitudinal rate of change of the ship's cross-sectional area~\cite{gmdsa_r66,gmdsa_r68}, and this aspect can be well captured by GMs, especially those of higher order. In fact, an early derivation for the evaluation of $C_w$ for slender ships, known as Vosser's integral, reveals explicit dependence on the longitudinal derivative of the cross-sectional area~\cite{gmdsa_r68}, i.e.,  $S^\prime(x)=\frac{\text{d}}{\text{d}x}S(x)$ where $S(x)=\int_{\Omega(x)}\text{d}y\text{d}z$ is the cross-sectional area, and $\Omega(x)$ denotes the cross-section of a ship hull at the longitudinal position $x$. 

\noindent Let now $m_p=\int_o^Lx^pS^\prime(x)\text{d}x$ be the $p-$th order moment of $S^\prime(x)$ with $x=0$ and $x=L$ corresponding to the stern and bow tips of the hull, respectively. Assuming that $S(0)=S(L)=0$ we get:
    \begin{equation}\label{gmdsa_e29}
    m^p = -p\int_0^L x^{p-1} S(x)\text{d}x = -p \int_0^L\int_{\Omega(x)}x^{p-1}\text{d}x\text{d}y\text{d}z,
    \end{equation}
    \noindent which leads to 
    \begin{equation}\label{gmdsa_e30}
    m^p = -pM^{p-1,0,0},
    \end{equation}
    \noindent where $M^{p-1,0,0}$ is a component of the hull's GMs vector of order $s=p+q+r=p-1$; see Eq. \eqref{gmdsa_e2}. Thus, $p-$order 1D moments of $S^\prime(x)$ are directly linked to $(p-1)-$order 3D longitudinal GMs of the hull. These physics-informed moments are included in the set of GMs used for building the SST we use for training ShipHullGAN.

Obviously, one cannot expect that every physical QoI of integral character is strongly connected with the GMs of the ship shape. Therefore,  the usage of GMs can only cover some physics-informed features. For example, viscous-pressure resistance is expressed as an integral over the wetted surface of the body; nevertheless, it depends on local properties of the surface, such as smoothness and curvature, which can act as turbulence generators by triggering flow separation. However, even if there is no strong connection between GMs and physics quantities under consideration, the usage of the former can still provide high-level intrinsic geometric information of the shape's geometry, which is imperative for extracting efficient features with enhanced diversity and geometric validity.

\subsubsection{Augmenting the final dataset with GMIs}
The employed SST for the training dataset instances incorporates GMIs of up to $s=4$th order with $n_M=35$ components. As an example, the GMIs for DTC, Series-60 and S-175 hulls depicted in Fig.~\ref{shipgan_f1} are reported in Table~\ref{table_1}. Higher-order GMIs can be utilised, but as the order increases, the GMIs become more susceptible to noise, necessitating more careful handling. This also increases the potential for numerical inaccuracies and computational issues. Luckily, as demonstrated in~\cite{khan2022geometric,khan2022shape}, $4$th order GMIs are sufficient for capturing geometric features and the associated physics ($C_w$) in ship design. 

    \begin{table}[htb!]
    \small
    \centering
    \caption{Geometric moment invariants up to $4\text{th}-$order evaluated for the DTC, Series-60 and S-175 hulls in Fig. \ref{shipgan_f2}.}
    \begin{tabular}{llllllllll}
    \toprule
     &$MI_{0,0,0}$ & $MI_{1,0,0}$ & $MI_{0,1,0}$ & $MI_{0,0,1}$ & $MI_{2,0,0}$ & $MI_{0,2,0}$ & $MI_{0,0,2}$ & $MI_{1,1,0}$ & $MI_{1,0,1}$  \\
    \midrule
    DTC&1.00 & 0.00 & 0.00 & 0.00 & 1.39 & 3.25E-02 & 1.24E-02 & 0.00 & -1.62E-02  \\
    S-175 & 1.00 & 0.00 & 0.00 & 0.00 &  1.45  & 3.86E-02 & 1.01E-02 & 0.00  & -9.37E-03\\
    Series-60&1.00 & 0.00 & 0.00 & 0.00 & 1.26  & 3.12E-02  & 1.36E-02  & 0.00  & -9.62E-03\\
    \midrule
    &$MI_{0,1,1}$ & $MI_{0,0,3}$ & $MI_{0,1,2}$ & $MI_{0,2,1}$ & $MI_{0,3,0}$& $MI_{1,0,2}$&$MI_{1,1,1}$ & $MI_{1,2,0}$ & $MI_{2,0,1}$     \\
    \midrule
     DTC& 0.00 & -2.75E-04 & 0.00 & 4.05E-04 & 0.00 & 5.65E-04 & 0.00 & -2.72E-03 & 3.13E-02  \\
     S-175 & 0.00  & -1.64E-04 & 0.00  & 4.91E-04 & 0.00  & 1.90E-05 & 0.00  & -3.38E-03 & 1.86E-02\\
     Series-60& 0.00  & -1.73E-04  & 0.00  & 2.40E-04  & 0.00  & 1.01E-05  & 0.00  & -3.11E-04  & 1.99E-02\\
     \midrule
     &$MI_{2,1,0}$ & $MI_{3,0,0}$  & $MI_{0,0,4}$ & $MI_{0,1,3}$ & $MI_{0,2,2}$ & $MI_{0,3,1}$ & $MI_{0,4,0}$ &$MI_{1,0,3}$ &$MI_{1,1,2}$  \\
     \midrule
      DTC&0.00 & -9.01E-02 & 2.90E-04 & 0.00 & 3.71E-04 & 0.00 & 2.07E-03 & -3.27E-04 & 0.00  \\
      S-175 &   0.00  & 1.51E-01 & 1.88E-04 & 0.00  & 3.61E-04 & 0.00  & 3.27E-03 & -1.69E-04 & 0.00\\
      Series-60   & 0.00  & -3.56E-02  & 3.36E-04  & 0.00  & 4.07E-04  & 0.00  & 1.96E-03  & -2.23E-04  & 0.00 \\
    \midrule
      &$MI_{1,2,1}$& $MI_{1,3,0}$ & $MI_{2,0,2}$&$MI_{2,1,1}$ & $MI_{2,2,0}$ & $MI_{3,0,1}$ & $MI_{3,1,0}$ & $MI_{4,0,0}$\\
    \midrule
    DTC       & -7.35E-04 & 0.00 & 1.65E-02 & 0.00 & 3.84E-02 & -5.77E-02 & 0.00 & 3.91\\
    S-175     & -3.35E-04 & 0.00  & 1.46E-02 & 0.00  & 3.03E-02 & -4.59E-02 & 0.00  & 4.86 \\
    Series-60 & -3.41E-04  & 0.00  & 1.70E-02  & 0.00  & 2.74E-02  & -3.43E-02  & 0.00  & 3.24\\
    \bottomrule
    \end{tabular}
    \label{table_1}
    \end{table}

\begin{figure}[htb!]
    \centering
    \includegraphics[width=01\textwidth]{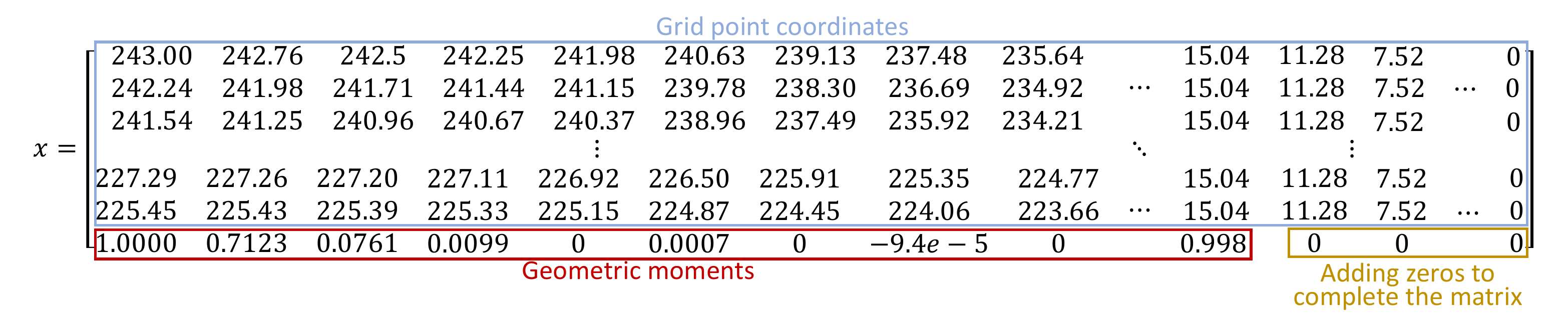}
    \caption{Structure of a matrix containing coordinates of the grid points and GMIs of a design in the training dataset. }
    \label{shipgan_f18}
    \end{figure}

    Once GMIs of a design is obtained, all of its 35 components are added to the last row of the matrix containing the grid point coordinates of that design; see Fig. \ref{shipgan_f18}. Afterwards, zeros are added in the remaining 22 elements to complete the $[25\times 57]$ dimensional matrix. Such matrices containing $x$, $y$ and $z$ coordinates, along with the corresponding GMIs, constitute the SST for each design in the training dataset and are provided as input when training the ShipHullGAN model. The rich representation of training designs resulting from the proposed shape encoding approach and their GMs generate high-resolution design, as they can learn more complex and hierarchical relationships between the initial distribution and the generated design.

\subsection{Enhancing shape diversity} 
Inadequately trained GANs fail to model the entire design space corresponding to the training data, with the resulting generator producing designs only in \say{neighbourhoods} around design clusters in the dataset. This results in a non-uniform coverage of the design space and lack of diversity~\cite{goodfellow2016nips}. This problem is quite prominent when the training dataset is composed of designs with different classes, as in our present case. Generators with these limitations can be easily analysed by examining groups of similar or identical generated designs and noting any unwanted clustering behaviour.

To reduce the likelihood of such an outcome, we introduce a space-filling criterion that enables ShipHullGAN to map latent features on the entire training space and enhance diversity. This criterion is implemented using the Audze and Eglais approach as presented in~\cite{intra_r46}. This approach achieves uniformity by mimicking the process of reaching minimum potential energy in physical systems. More specifically, it follows a physical analogy to the repulsive forces exerted by molecules, designs, in our case, in space. The molecules are in equilibrium when minimum potential energy is reached, which subsequently guarantees uniform distribution over the entire design space. Assuming the existence of several design subclasses in the design space, the criterion for $m$ designs from the generator is evaluated as
    
\begin{equation}\label{E24}
\mathcal{S}=\sum_{i=1}^{m-1}\sum_{j=i+1}^{m}{\frac{1}{||\mathbf{x}_{j}-\mathbf{x}_i||_2^2}}.
\end{equation}

\noindent where, $\mathbf{x}_i$ and $\mathbf{x}_j$ constiture a pair of generated designs. Minimisation of $\mathcal{S}$ favours their uniform distribution of designs over the entire design space. For more details on space-filling, the interested reader should refer to~\cite{khan2018generative}.

\subsection{Loss function}
The space-filling term in Eq. \eqref{E24} is then added to the original loss function of the GAN (see Eq. \eqref{GANlossFun}), resulting in the new augmented loss function below:

\begin{equation}
     \min_G\max_D\mathcal{L}_{adv} (D,G) + \Gamma_G~\mathcal{S},
\end{equation}

\noindent where $\Gamma_G$ controls the contribution of the space-filling term. Typically, at the initial phases of GAN training, the generation of invalid/unrealistic designs is more probable; therefore, at this stage, we set $\Gamma_G$ equal to 0 and increase it gradually during training so that ShipHullGAN focuses firstly on learning to generate realistic designs at the early stages and then space-filling criterion kicks in to uniformly generate designs in the design space. During training, $\Gamma_G$ is set on an escalating schedule proposed in \cite{ssdr_r26}, which is formulated as 

\begin{equation}
    \Gamma_G = \Gamma_G^{\prime}\left(\frac{t}{T}\right)^p,
\end{equation}

\noindent where $\Gamma_G^{\prime}$ is the value of $\Gamma_G$ at the end of training, $t$ is the current training step, $T$ is the total number of training steps, and $p$ is a factor controlling the steepness of the escalation.

\subsection{Model architecture details and training considerations}
In this last part of section~\ref{sec:ShipHullGAN}, we discuss some technical \& architectural details about the ShipHullGAN model's components, generator and discriminator, along with additional considerations for its input that will enable appropriate training of the proposed GAN model.

\subsubsection{Architecture of generator and discriminator}
\begin{figure}[htb!]
    \centering
    \includegraphics[width=01\textwidth]{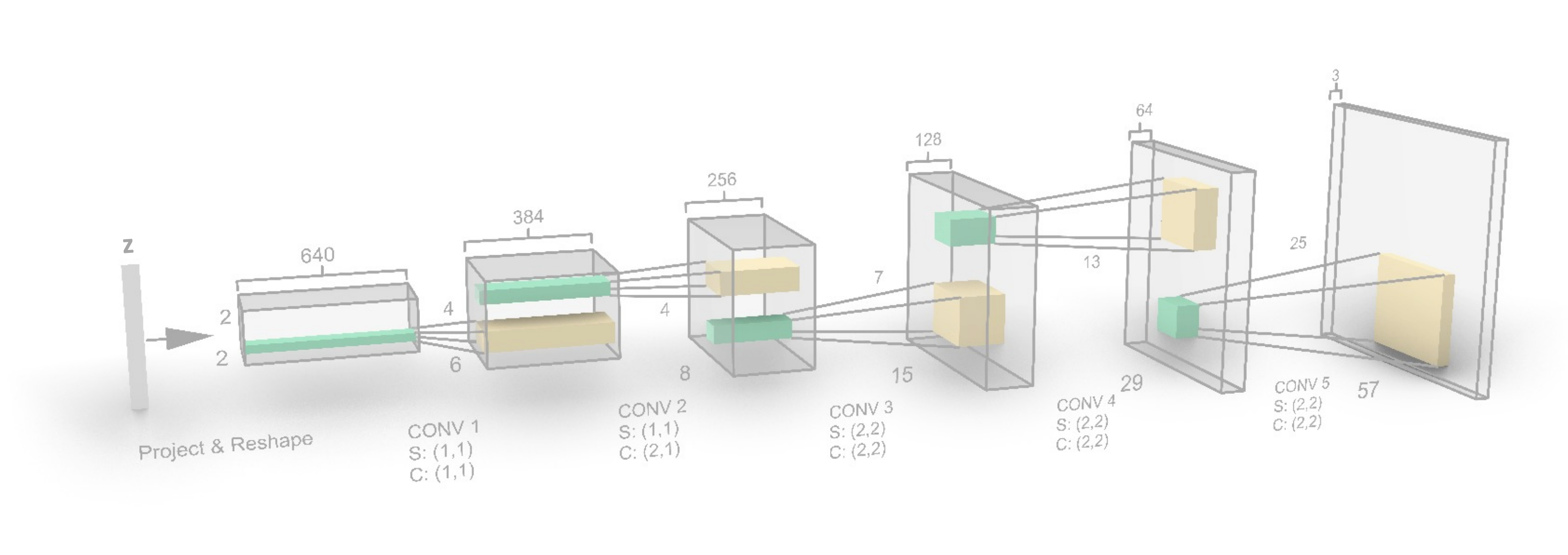}
    \caption{Convolutional architecture of the generator used in ShipHullGAN. }
    \label{shipgan_f14}
    \end{figure}

As mentioned at the beginning of the section, the generator, $G$, and discriminator, $D$, are materialised via deep convolutional neural networks whose structure is shown in Fig. \ref{shipgan_f14}. The discriminative network, $D$, consists of 6 convolutional layers and one input layer, which takes three $[25\times57]$ matrices of grid points ($x$, $y$ and $z$ coordinates) augmented with 4th order GMIs. A dropout layer, with a dropout probability of 0.5, succeeds the input layer to prevent over-fitting on the training data. This layer acts as a mask that randomly nullifies the contribution of some neurons toward the next layer. An activation layer follows each convolutional layer with a leaky rectified linear activation function (ReLU). The last convolutional layer uses a sigmoid activation function that calculates the probability of a design being fake or real. For the second, fourth and fifth convolutional layers, batch normalisation is applied before the ReLU layer. The discriminator typically reduces data dimensions when assessing whether a design is real or fake in an operation that resembles downsampling when dealing with images. This downsampling in $D$ is performed with strides of different padding sizes instead of the common pooling layer, as strides tend to improve the accuracy and stability of the model; see~\cite{li2020efficient}. 

The generator, $G$, is the transpose of $D$ and comprises 5 transposed convolutional layers, along with an input, projection and reshape layer. The input layer takes a randomly sampled $\mathbf{z}$ from a given distribution and feeds it to the \say{project and reshape} layer. Apart from the last layer, each convolutional layer is followed by batch normalisation and ReLU. The last convolutional layer of $G$ has an activation layer with a hyperbolic tangent function to ensure an output value between -1 and 1, generating the normalised $[25\times57]$ matrices corresponding to our SST.

This architecture resulted from systematic experimentation described in \S\ref{expShipGAN} and secures an adequately stable and smooth training procedure; additional details about the selection process and possible enhancements are given in \S\ref{expShipGAN}. Model training is performed with the Adam gradient descent algorithm on a PC with dual 24-core 2.7GHz Intel\textsuperscript{\tiny\textregistered} Xeon\textsuperscript{\tiny\textregistered} 6 Gold 6226 CPU, NVIDIA Quadro RTX 6000 GPU and 128GB of memory, using the following settings: number of epochs = 500; minimum batch size = 128, learning rate = 0.0002 and gradient decay factor = 0.5. Generator and discriminator networks employ 9.7 and 9.6 million learnable parameters, respectively.

\subsubsection{Size of the input feature vector $\mathbf{z}$}
Unlike other techniques, such as principal component analysis (PCA) and others, the determination of the latent vector's ($\mathbf{z}$) size can be challenging in GANs. The deep convolutional neural networks utilised for both generator and discriminator allow our model to generate samples of higher quality, as they capture the hierarchical features of the target data distribution. This not only stabilises the training process but also helps to avoid mode collapse. However, an inappropriate size for $\mathbf{z}$ can still lead to mode collapse, with the generator mapping multiple $\mathbf{z}$ vectors to the same output~\cite{goodfellow2016nips}. Especially when $\mathbf{z}$ is small, the possibility of the generator's failure to cover the entire training dataset distribution increases, and it may produce many invalid designs and/or designs with minimal diversity. Obviously, a larger $\mathbf{z}$ may resolve this, but not without cost since large vectors correspond to high-dimensional design spaces when performing shape optimisation, which increases the computational complexity of the entire simulation-driven design pipeline \cite{khan2022shape}. Therefore, for estimating a sufficient but not redundant size of $\mathbf{z}$, we perform PCA and use the number of eigenvalues required for achieving a target variance as a reasonable estimation of the initial size of $\mathbf{z}$. 

    \begin{figure}[htb!]
    \centering
    \includegraphics[width=0.45\textwidth]{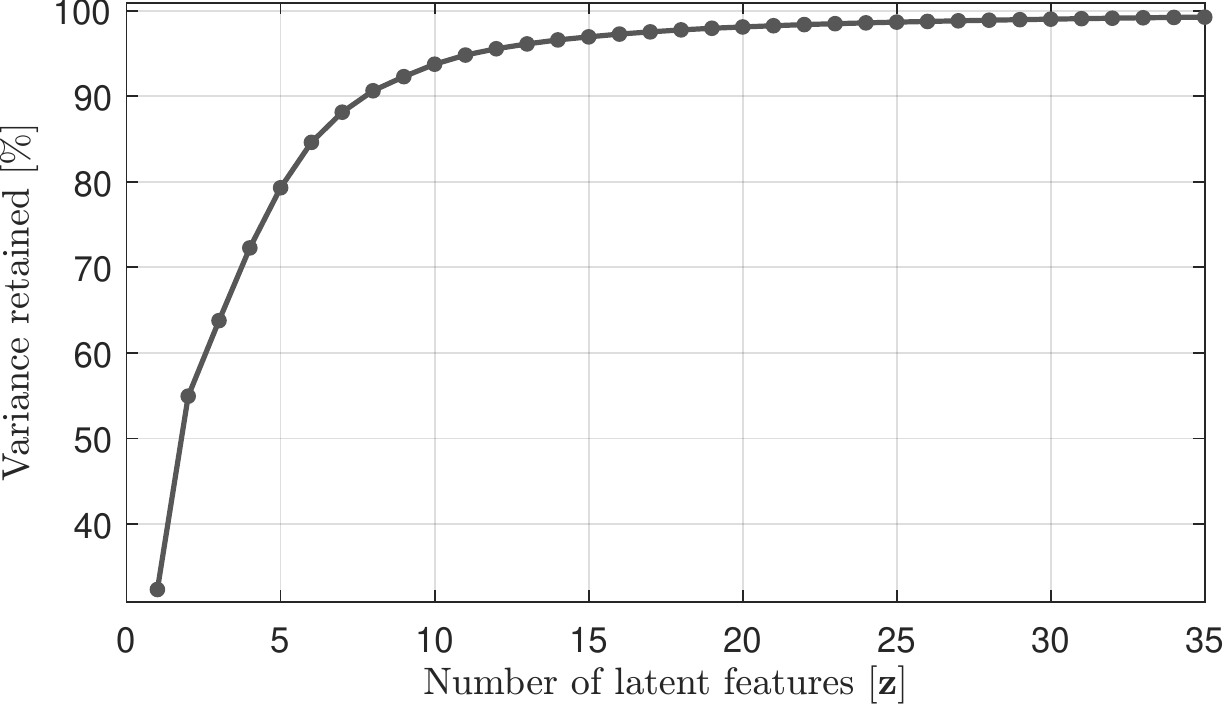}
    \caption{Percentage of variance retained versus size of $\mathbf{z}$.}
    \label{shipgan_f15}
    \end{figure}

As it can be easily seen from Fig.~\ref{shipgan_f15}, 30 latent features in $\mathbf{z}$ can capture 99\% of geometric variance. We, therefore, set the initial size of $\mathbf{z}$ to 30 and then reduce it iteratively while measuring the diversity, novelty, and maximum mean discrepancy (MMD)~\cite{li2020efficient} of generated designs. The variety and novelty are estimated with the sparseness at the centre (SC)~\cite{brown2019quantifying} and the novelty score described in~\cite{ssdr_r26}, respectively. The MMD metric is evaluated using Eq.~ \eqref{shipgan_eq_MMD} below, which measures the similarity between the distribution of designs in the training dataset and designs resulting from the generator. A high value of the MMD means that the generator cannot completely cover the design space in the training dataset, which may indicate a mode collapse issue. We may also note here that as GAN incorporates nonlinear layers, it should be able to capture the variability and nonlinearity in the training dataset with fewer latent variables compared to PCA. Thus, the initial size 30 can also be considered as an upper bound for the size of $\mathbf{z}$.

\begin{equation}\label{shipgan_eq_MMD}
    \mathrm{MMD} = \frac{1}{n^2}\sum_{i=1}^{n}\sum_{j=1}^{n}k\left(\mathbf{x}^i,\mathbf{x}^j\right)+\frac{1}{m^2}\sum_{i=1}^{m}\sum_{j=1}^{m}k\left(\mathbf{x}^i_{GAN},\mathbf{x}^j_{GAN}\right)-\frac{2}{nm}\sum_{i=1}^{n}\sum_{j=1}^{m}k\left(\mathbf{x}^i,\mathbf{x}^j_{GAN}\right),
\end{equation}

\noindent In the above equation, $\mathbf{x}$ and $\mathbf{x}_{GAN}$ correspond to designs in the training dataset and designs generated from the generator, respectively,  with $n$ and $m$ being the corresponding total numbers of the two sets of designs. Finally, $k$ is a radial kernel function defined as 

\begin{equation}
    k(\mathbf{x},\mathbf{y}) = \text{exp}\left(-\frac{||\mathbf{x}-\mathbf{y}||_2}{2\theta^2}\right),
\end{equation}

\noindent with $\theta=0.1$. 

We evaluate SC and novelty metrics using Eqs.~\eqref{shipgan_eq_D} and \eqref{shipgan_eq_novelty}, respectively. The SC measures the average distance of the centroidal design, $\mathbf{x}^{centroid}_{GAN}$, to the $m$ designs resulting from ShipHullGAN. In contrast, novelty evaluates how different newly generated designs are from the designs in the training dataset, $\mathcal{X}$. It is estimated first by finding the nearest distance between the $i$th new design, $\mathbf{x}_{GAN}^i$, and all $n$ designs in $\mathcal{X}$, and then by averaging all of those $m$ nearest distances. 

\begin{equation}\label{shipgan_eq_D}
    SC  = \frac{1}{m}\sum_{i=1}^m||\mathbf{x}^{centroid}_{GAN}-\mathbf{x}^i_{GAN}||_2
\end{equation}

\begin{equation}\label{shipgan_eq_novelty}
    Novelty = \frac{1}{m}\sum_{i=1}^m \min_{\mathbf{x}^j\in\mathcal{X}}||\mathbf{x}^i_{GAN}-\mathbf{x}^j||_2.
\end{equation}

\noindent  Here, $\mathbf{x}^j$ are the designs in the training dataset, $\mathcal{X}$.

    \begin{figure}[htb!]
    \centering
    \includegraphics[width=01\textwidth]{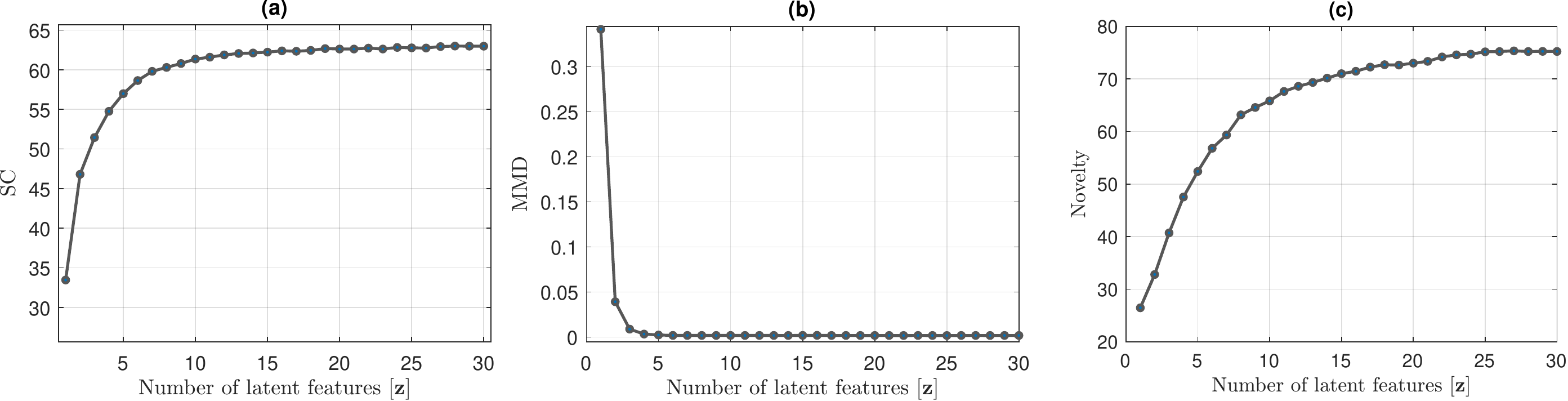}
    \caption{Plots depicting the value of (a) SC, (b) MMD and (c) novelty metrics evaluated using Eqs.~\eqref{shipgan_eq_D}, \eqref{shipgan_eq_MMD} and \eqref{shipgan_eq_novelty}, respectively, versus the number of employed latent features. }
    \label{shipgan_f19}
    \end{figure}

We analyse the influence of latent space dimensionality against these three metrics in Fig. \ref{shipgan_f19}. Higher values of SC and novelty generate diverse designs, while low values of MMD correspond to good coverage of the design space $\mathcal{X}$ by the generator. Fig.~\ref{shipgan_f19} clearly indicates that as the number of latent features increases, diversity and novelty increase approximately up to the number of 20 features and then tend to plateau. In contrast, the MMD reduces rapidly and reaches a sufficiently low value with 5 features. As mentioned earlier, The size of the feature vector $\mathbf{z}$ plays a crucial role in the output of the generator. A small size may result in irregular or identical geometries, while a large size leads to a high-dimensional design space for optimisation, as previously discussed in literature \cite{li2020efficient,khan2022shape}. Therefore, in view of these points, analysis of results in Fig.~\ref{shipgan_f19} indicate that 20 features is a well-balanced selection for the size of $\mathbf{z}$, and as it will be demonstrated in the subsequent section, a generator trained with 20 features produces valid and physically-plausible designs.

\section{Experiments: Design synthesis and optimisation} 
This section presents the process and experimentation results used to validate the appropriateness and efficiency of the proposed model.
\label{expShipGAN}
\subsection{Design reconstruction}\label{designRecinsSec}
\begin{figure}[htb!]
    \centering
    \includegraphics[width=01\textwidth]{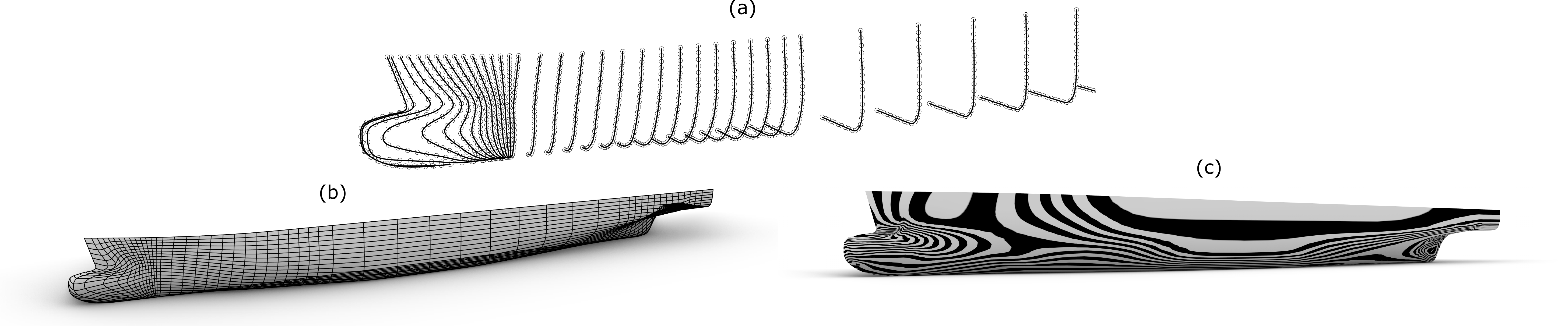}
    \caption{(a) Interpolation of points of CSs using cubic NURBS curves. (b) Construction of NURBS surfaces interpolating the curves with a loft operation. (c) Inspection of hull surface fairness using isophotes mapping analysis.}
    \label{shipgan_f13}
    \end{figure}

After the training process has been completed, we use the generator of the trained model as a parametric modeller with 20 parameters ranging between -1 and 1, generating design in a 20-dimensional subspace, $\mathcal{Z}$. For an input vector $\mathbf{z}$ sampled from $\mathcal{Z}$, the generator produces three $[25\times57]$ matrices corresponding to the $x$, $y$ and $z$ coordinates of grid points of a new design. Recall that the last row in these matrices corresponds GMIs; therefore, we remove this row from all three matrices to construct the final shape. The shape reconstruction using a NURBS surface of the new design is generated by first fitting a NURBS curve to the points of each cross section (CS); see Fig.~ \ref{shipgan_f13}(a). Then, the 3D surface representation is created by interpolating the reconstructed CSs with a bicubic NURBS surface using a skinning scheme (a.k.a. loft operation) as shown in Fig.~\ref{shipgan_f13}(b). The resulting surface is smooth and fair with sufficient continuity, as indicated by using an isophotes mapping analysis (zebra stripes) on the reconstructed hull surface shown in Fig.~\ref{shipgan_f13}(b)). The smooth transition of the zebra stripes on the surface indicates a smooth and fair hull surface of $C^2$ continuity. 

\begin{figure}[htb!]
    \centering
    \includegraphics[width=01\textwidth]{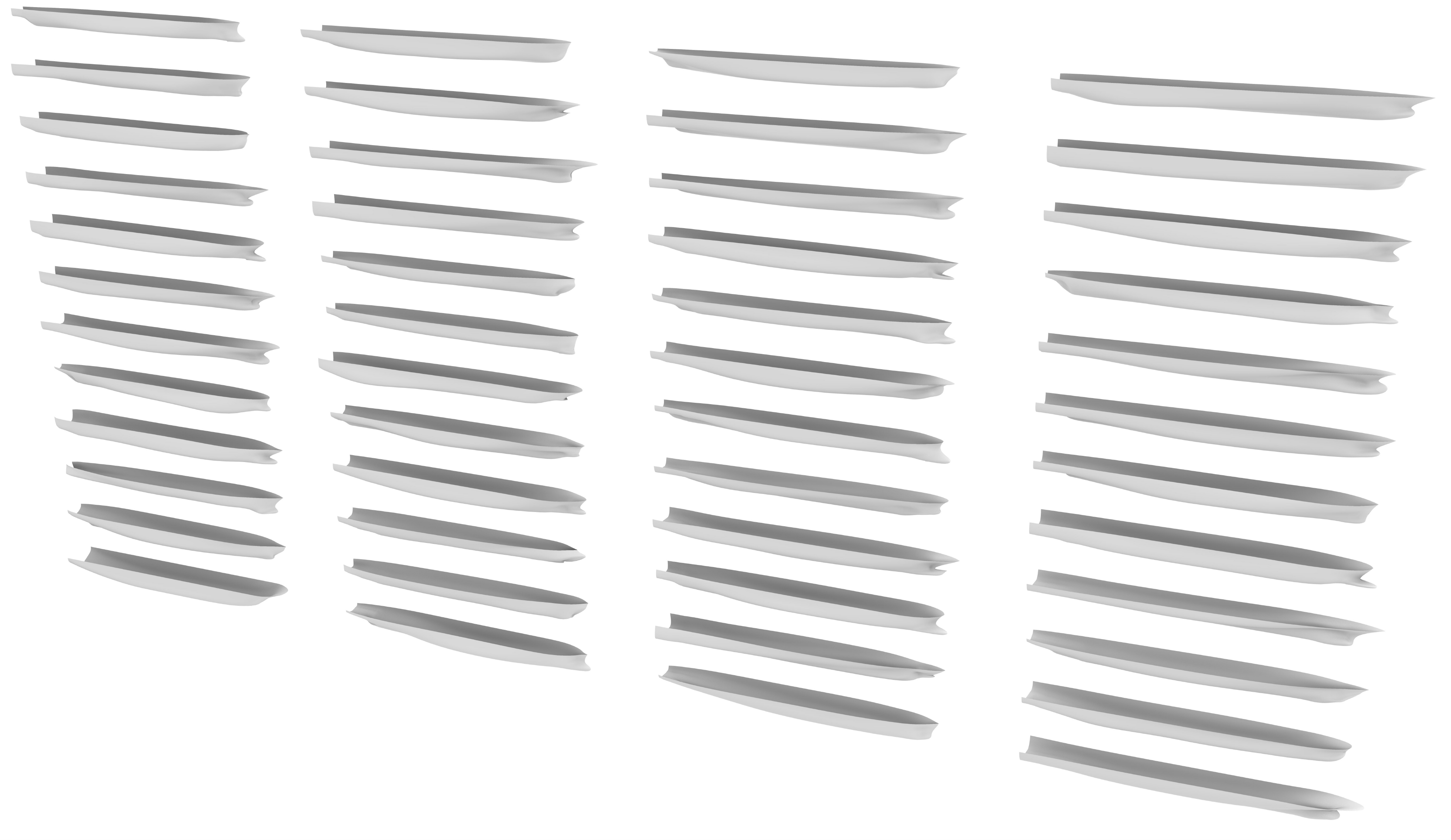}
    \caption{Design variations created with ShipHullGAN. Randomly sampled designs from $\mathcal{Z}$ and design variations resulting from changing each of the variables in $\mathbf{z}$ can be visualised at \url{https://youtu.be/ZIfmAs5-qFw} and \url{https://youtu.be/avlq0FxZP-s}, respectively.}
    \label{shipgan_f8}
    \end{figure}

\begin{figure}[htb!]
    \centering
    \includegraphics[width=01\textwidth]{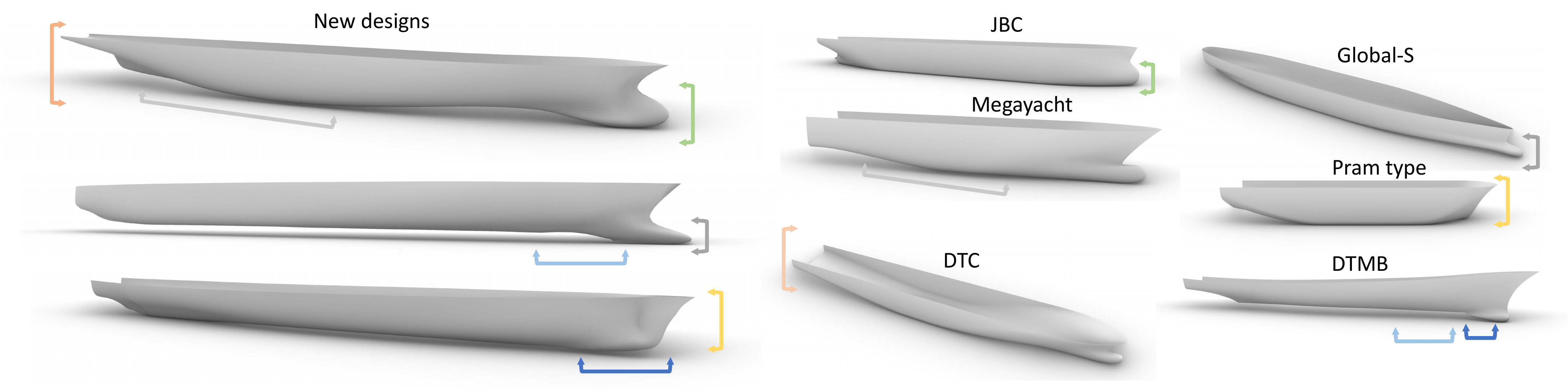}
    \caption{Examples of newly generated designs using ShipHullGAN adopting features from parent designs in Fig.~\ref{shipgan_f2}.}
    \label{shipgan_f23}
    \end{figure}

Indicative variations of the ship hulls generated using the ShipHullGAN model are shown in Fig.~\ref{shipgan_f8}. From a visual inspection of these designs, a designer can easily conclude that these designs are physically valid and plausible with distinct geometric features and characteristics. One can also quickly identify augmented features from the designs in the training dataset on several of the generated designs. In Fig.~\ref{shipgan_f23}, we depict three generated hulls from the ShipHullGAN model and the correspondence of their features to existing hulls. For example, the new design on the top right corner of Fig.~\ref{shipgan_f23} adopts features in the bow (green arrows), aft (grey arrows), and stern (orange arrows) regions, resembling JBC, Megayacht and DTC parent hull features\footnote{see also Fig.~\ref{shipgan_f2}}, respectively. This supports our claim that the proposed generic parametric model can generate hulls with diverse features from completely different ship hull types, which is one of the features existing parametric modelling approaches in hull design largely lack. 

\subsection{Design validity and diversity}
The geometric validity of designs resulting from the model is partially tested by searching for designs with self-intersecting geometries. We randomly sampled 30,000 designs over ten runs and searched for self-intersecting geometries. Interestingly enough, no self-intersections were found in any of the 300,000 tested designs. This is a strong indication that the ShipHullGAN model is robust and efficient, and these properties are attributed to its convolutional architecture, reliable training and inclusion of GMIs in the SST. 

\begin{figure}[htb!]
    \centering
    \includegraphics[width=01\textwidth]{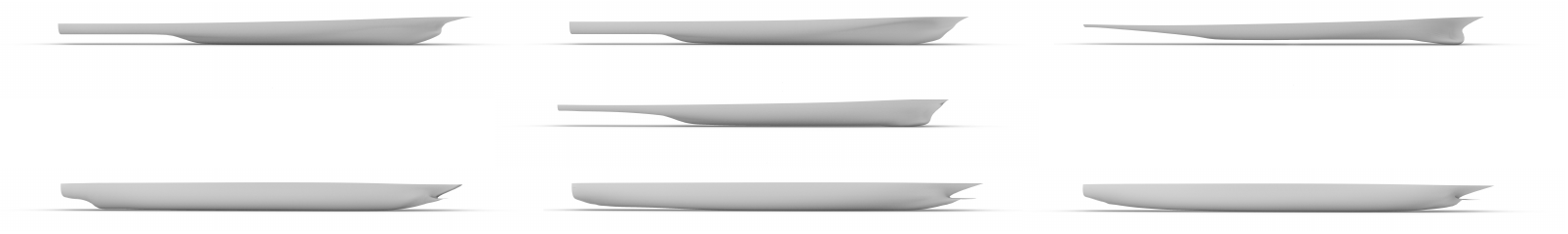}
    \caption{Example of implausible designs.}
    \label{shipgan_f10}
    \end{figure}

However, even though no self-intersecting geometries were detected, some of the ShipHullGAN-generate designs may be implausible from a practical point of view. Examples of such designs are shown in Fig.~\ref{shipgan_f10}. Nevertheless, the possibility of receiving such designs is rather low as a visual inspection of large numbers of randomly sampled designs resulted in less than 1 out of 70 instances with questionable designs. However, such designs can be eliminated by setting appropriate design constraints and/or employing the physical solver to rule out such designs during design optimisation. 

    \begin{figure}[htb!]
    \centering 
    \includegraphics[width=0.7\textwidth]{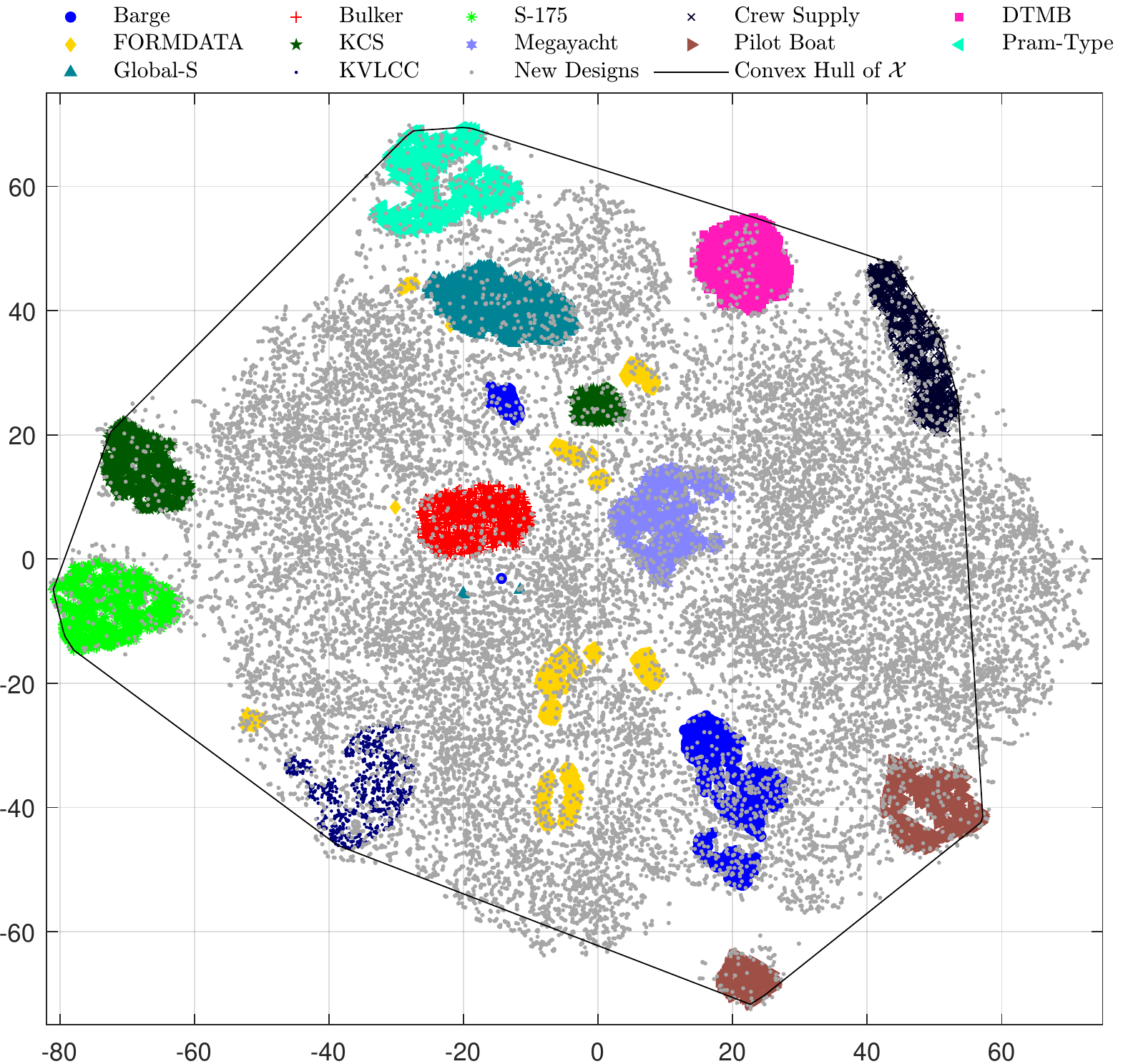}
    \caption{t-SEN plot of some design in the training data and newly generated designs from the ShipHullGAN model.}
    \label{shipgan_f11}
    \end{figure}

We also use t-distributed stochastic neighbour embedding (t-SNE)~\cite{van2008visualizing} to analyse further the diversity and its ability to cover the design space of the training data. t-SNE is a statistical method for visualising high-dimensional data by giving each data point a location in a 2D or 3D map and can provide some indication of the distribution of designs. From Fig.~\ref{shipgan_f11}, it can be seen that newly generated designs cover well the entire convex hull enclosing the designs in the training dataset. It should be noted that the topology of the t-SNE plot, more precisely the distance between the cluster, their size and orientations, may not have any physical meaning; therefore, in the present case, it's mainly used to visualise the distribution of generated designs within the training space. Moreover, as can be seen in the same figure, some of the new designs reside out of the convex hull, which according to \cite{ssdr_r26}, further indicates the ability of the generator to create novel designs. In summary, these results demonstrate that the parametric modeller resulting from ShipHullGAN is able to generate

\begin{enumerate}
    \item designs similar to the training dataset (new designs overlap the existing ones),
    \item designs with augmented features from different classes of design in the training dataset (new designs between the clusters), and
    \item completely novel designs (new designs outside the convex hull).
\end{enumerate}

\subsubsection{Comparison with GAN}
\begin{figure}[htb!]
    \centering
    \includegraphics[width=0.75\textwidth]{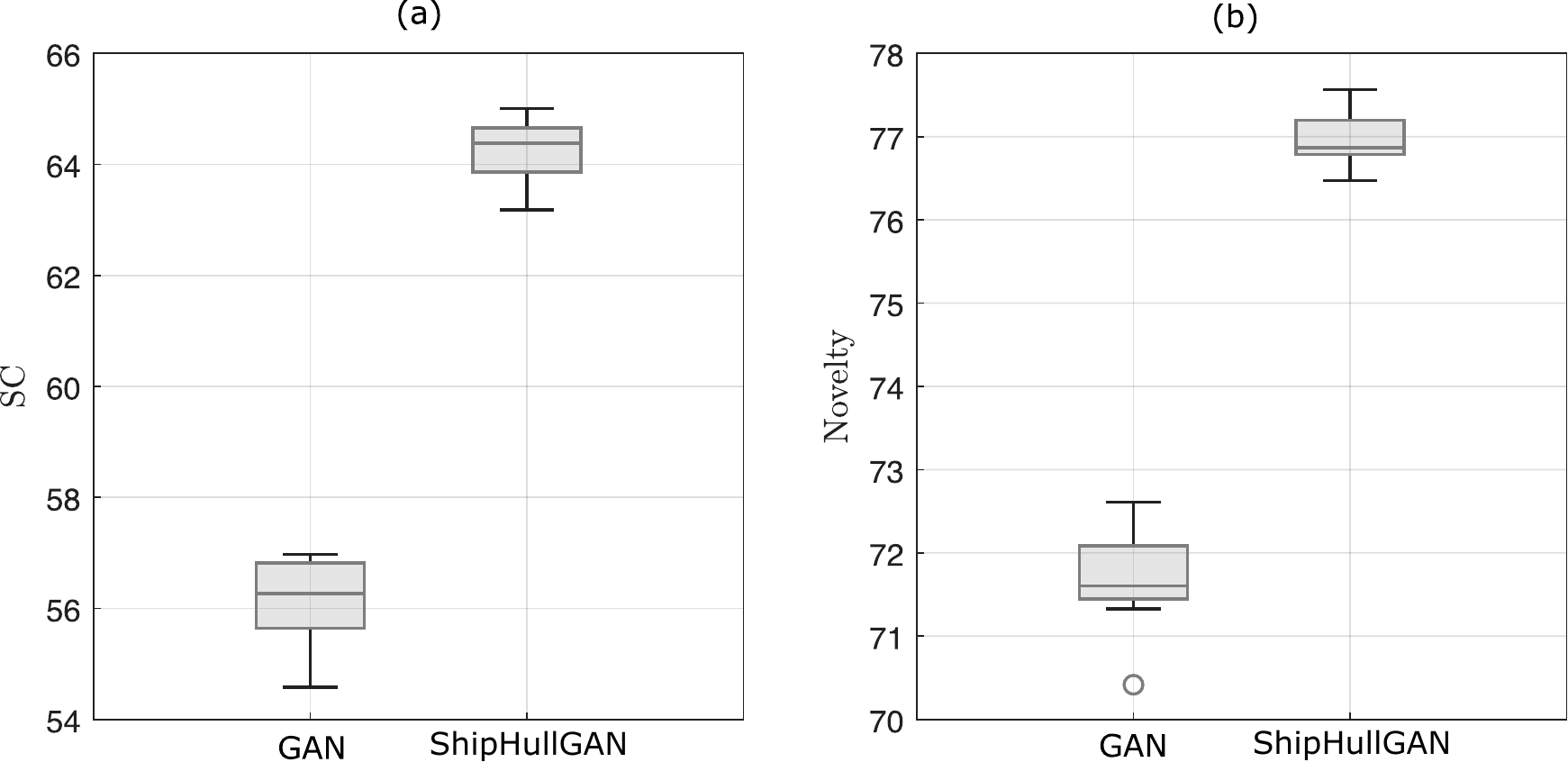}
    \caption{(a) Diversity and (b) novelty of designs created with the generator of GAN and ShipHullGAN.}
    \label{shipgan_f20}
    \end{figure}

We finally compare ShipHullGAN with a GAN model trained with the exact same settings and architecture as ShipHullGAN but without space-filling and GMIs components to highlight their respective impact. We first evaluate the SC metric for both models, using 30,000 randomly sampled designs over ten runs (300,000 designs in total). The results of this experiment are shown in Fig.~\ref{shipgan_f20}. It can be easily seen that the ShipHullGAN model shows significantly higher diversity and novelty compared to the GAN. We also conducted a t-test to see if there exists a significant difference between the diversity values. The $p-$values resulting from this test are $3.7354E-09$ and $2.1315E-09$, respectively, which are lower than 0.05, indicating a significant difference.  

\begin{figure}[htb!]
    \centering
    \includegraphics[width=01\textwidth]{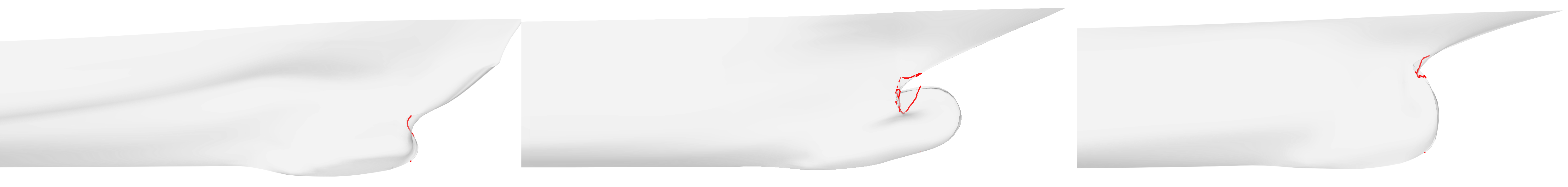}
    \caption{Examples of invalid (self-intersecting) designs resulted from the GAN model. The red curve indicates the regions of intersection.}
    \label{shipgan_f22}
    \end{figure}

    Furthermore, we also analysed the ability of GAN to produce valid designs, i.e., designs with non-self-intersecting surfaces, by once again sampling 30,000 designs over ten runs and averaging the number of invalid over valid designs. As discussed earlier, for a similar test, ShipHullGAN resulted in zero invalid designs; however, approximately 4.32\% of designs resulting from GAN were invalid. Although this difference is not so significant, it still demonstrates the capability of ShipHullGAN to produce valid geometries, mainly due to the usage of geometric moments in the SST. Moreover, most invalid designs resulting from GAN have self-intersecting surfaces near the bow of the hull, see Fig.~\ref{shipgan_f22}, which is a local feature. This shows that the GAN fails to capture the local features of the designs well due to the absence of rich information about the geometry, which in ShipHullGAN is given with the SST.

\subsection{Optimisation case studies}
The generic capabilities of the ShipHullGAN modeller can be exploited in different ways to support the designers throughout all three stages of the ship design; i) concept/preliminary design, ii) contract (full) design, and iii) detail (build) design, especially at the former two. In this section, we showcase two typical optimisation scenarios to help the readers envision how designers can use ShipHullGAN in practice in the preliminary and contract design phases.

As previously mentioned, ship hull optimisation is typically performed during the later stages of preliminary design or the contract phase for a \textit{specific parent design} that aligns with the given constraints and owner requirements. This is mainly due to the fact that existing parametric approaches can handle a single hull type and cannot aid the designer in the early phases of the preliminary design stage, where identification of a parent design and/or exploration of various innovative candidate solutions are essential. 

\subsubsection{Early-stage design optimisation}

With the aid of the generic parametric capabilities of ShipHullGAN, one can initiate design optimisation from the early preliminary design stages with a set of preliminary optimisation criteria, e.g., resistance for a range of speeds, and constraints, e.g., displacement, maximum breadth (e.g., to enable passing through the Panama channel) or maximum draft (e.g., for accessing specific ports). To showcase these capabilities, a simple optimisation problem is formulated aiming to explore the design space, $\mathcal{Z}$, for a container ship with a load-carrying capacity of 3600 TEU (Twenty-foot equivalent unit) and an oil tanker with 300,000 tons capacity with improved wave resistance coefficient $C_w$ by solving the optimisation problems in Eq.~\eqref{shipganOptEq1} and \eqref{shipganOptEq2} below, respectively. Note that $C_w=2R_w/({\rho}U^2S)$, where $R_w$ denotes the wave resistance, $\rho$ is the density of the seawater, $U$ is the ship's speed and, finally, $S$ is the wetted surface of the ship hull.

    \begin{equation}\label{shipganOptEq1}
    \begin{aligned}
    \textrm{Find } \mathbf{z^*}\in\mathbb{R}^{20} \quad & \textrm{such that} \\
    C_w(\mathbf{z^*}) = &  \min_{\mathbf{z} \in \mathcal{Z}} C_w(\mathbf{z}) \\
    \textrm{subject to:}\quad & \textrm{given cargo capacity (3600 TEU)};\\
    \quad & 51120.5m^3 \leq \text{Volume of displacement}~(\nabla) \leq 56501.6  m^3;\\
    \quad & 220.9m \leq \text{Length at waterline}~(L_{wl}) \leq 244.2m;\\
    \quad & 30.6m \leq \text{Beam at waterline}~(B_{wl}) \leq 33.8 m;\\
    \quad & 10.3m \leq \text{Draft}~(T) \leq 11.3m.
     \end{aligned}
    \end{equation}

    \begin{equation}\label{shipganOptEq2}
    \begin{aligned}
    \textrm{Find } \mathbf{z^*}\in\mathbb{R}^{20} \quad & \textrm{such that} \\
    C_w(\mathbf{z^*}) = &  \min_{\mathbf{z} \in \mathcal{Z}} C_w(\mathbf{z}) \\
    \textrm{subject to:}\quad & \textrm{given cargo capacity (300,000 tons)};\\
    \quad & 298723.8m^3 \leq \nabla \leq 330168.5m^3\\
    \quad & 309.2 \leq L_{wl} \leq 341.8m;\\
    \quad & 30.6m \leq B_{wl} \leq 33.8 m;\\
    \quad & 19.8 \leq T \leq 21.8m.
     \end{aligned}
    \end{equation}

    \begin{figure}[htb!]
    \centering
    \includegraphics[width=01\textwidth]{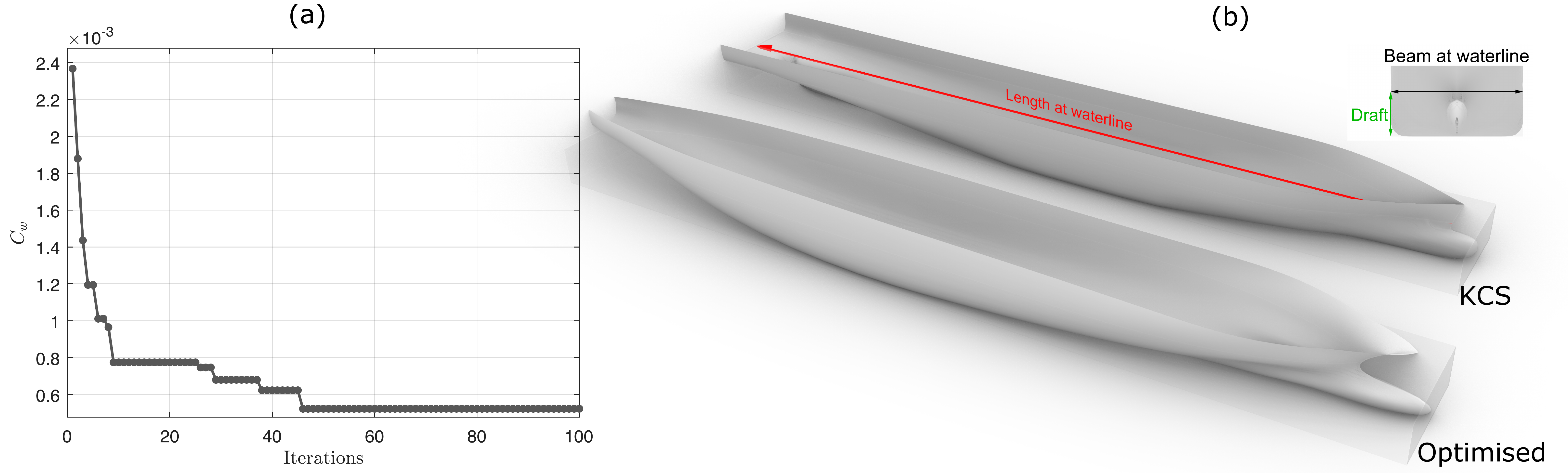}
    \caption{(a) Convergence plot of $C_w$ during the first 100 optimisation iterations. (b) 3D surfaces of the KCS and the ShipHullGAN-optimised hull with the same cargo capacity.}
    \label{shipgan_f17}
    \end{figure}

    \noindent The constraints in Eqs. \eqref{shipganOptEq1} and \eqref{shipganOptEq2} are set to derive an optimised design comparable to the KCS and KVLCC2 hulls shown in Fig.~\ref{shipgan_f2}. The KCS is the well-known 3600 TEU KRISO container ship designed by the Maritime and Ocean Engineering Research Institute (MOERI), while the KVLCC2 (KRISO Very Large Crude Carrier) represents a typical 300,000 tons tanker hull form which has been the subject of several experimental and computational studies in the pertinent literature.

\begin{figure}[htb!]
    \centering
    \includegraphics[width=01\textwidth]{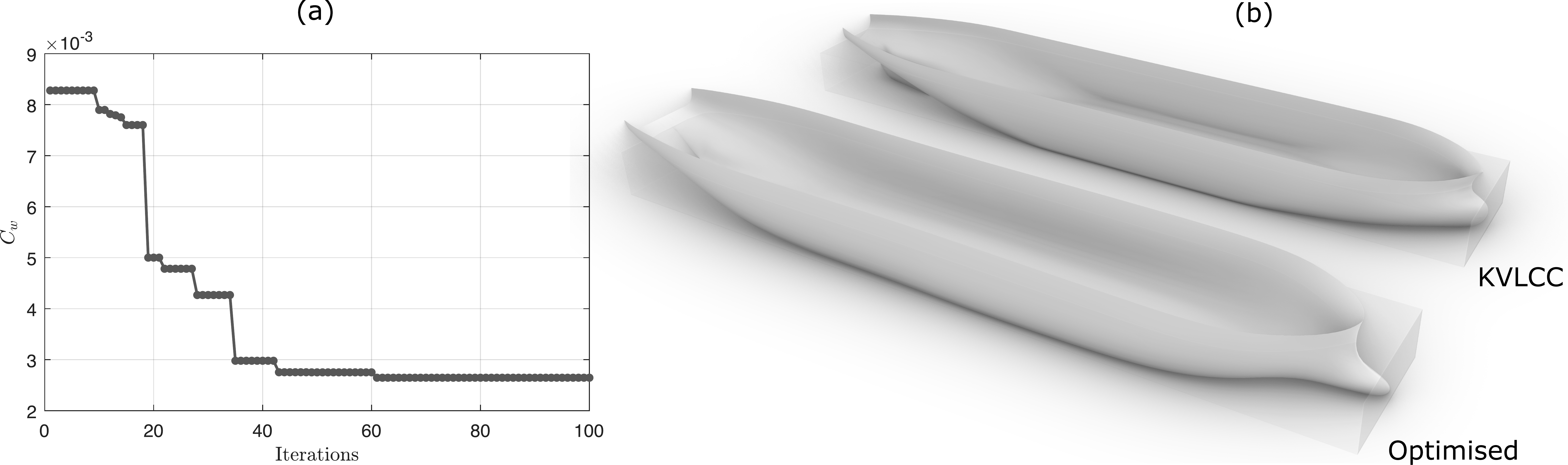}
    \caption{(a) Convergence plot of $C_w$ during the first 100 optimisation iterations. (b) 3D surfaces of the KVLCC hull and the ShipHullGAN-optimised hull with the same cargo capacity.}
    \label{shipgan_f24}
    \end{figure}

    The optimisation problems above are solved using Jaya Algorithm (JA), a simple yet efficient optimiser; see more details in \cite{intra_r64}. Hydrodynamic simulations for evaluating $C_w$ are performed using a software package based on linear potential flow theory using Dawson (double-model) linearisation, with details of the employed formulation, the numerical implementation, and its validation appearing in \cite{bassanini1994wave}. As a result of using simple Rankine sources, the computational domain consists of a part of the undisturbed free surface, extending 1$Lpp$ upstream,  3$Lpp$ downstream, and 1.5$Lpp$ sideways, with $Lpp$ denoting the length between perpendiculars for the assessed ship hull. A total of $[20 \times70]$ grid points are used for the undisturbed free surface, whereas  $[50\times180]$ grid points are used for the hull discretisation with the simulation being performed at a Froude number $F_r$ equal to $F_r=U/\sqrt{gL}=0.28$, where $g$ is the acceleration due to gravity, and $L$ is the ship's length.
    Furthermore, as JA employs a stochastic approach, results may slightly differ in each run; therefore, three runs are performed and averaged results are presented in this work. Figures~\ref{shipgan_f17}(a) and \ref{shipgan_f24}(a) display the convergence graph of $C_w$ over the first 100 iterations of the best of three runs; a total of 500 iterations is performed in each run. The optimised designs obtained in these cases, along with original KCS and KVLCC2 geometries, are depicted in Figs.~\ref{shipgan_f17}(b) and \ref{shipgan_f24}(b). 

The optimised designs in Figs.~\ref{shipgan_f17}(b) and \ref{shipgan_f24}(b) achieve $C_w$ values of 5.932E-04 and 2.646E-03, respectively, and comply with all the design constraints set in  Eqs. \eqref{shipganOptEq1} and \eqref{shipganOptEq2}. Table~\ref{table_2} summarises the results for the optimisation examples performed in this section. The achieved $C_w$ values are lower than the corresponding values of the parent KCS and KVLCC2 hulls, which are calculated at 2.477E-03 and 6.810E-03, respectively. As one may easily observe, the reported improvement is high, but this can be justified by taking into account a number of issues related to the optimisation setting and the limitations of the solver:
\begin{enumerate}
\item The obtained optimised designs differ significantly in shape from the corresponding KCS and KVLCC2 designs, as can be easily seen by observing the stem and stern areas in Figs.~\ref{shipgan_f17}(b) and \ref{shipgan_f24}(b). These designs are not traditional variations of the parent ones but stem from a more global shape optimisation, which is commenced without a parent design and, as a result, enables significant improvements
\item In these case studies, the quantity of interest is the wave-making resistance coefficient. If we included the remaining parts of the resistance, a more moderate improvement would be observed; for example, the obtained optimised designs possess a larger wetted surface, increasing the frictional resistance component.
\item Although potential flow codes are fast and efficient and are commonly employed in the early stages of the hull design process for exploring the design space by comparing quickly many design alternatives, they may not provide reliable performance evaluation, primarily when the design under consideration is composed of unconventional features. Therefore, in the future, we aim to run large-scale optimisation by employing computationally intensive CFD solvers to properly handle the impact of viscosity on total resistance.
\end{enumerate}
Nevertheless, these results demonstrate the generic parametric capabilities of the ShipHullGAN modeller that, under different design considerations, it cannot only create different valid design geometries but also demonstrate its capacity for design optimisation.

\begin{table}[htb!]
    \small
    \centering
    \caption{Main particulars and $C_w$ of KCS, KVLCC and Crew Supply vessel hulls and the optimised designs in the Figs.~\ref{shipgan_f17}, \ref{shipgan_f24} and \ref{shipgan_f25}.}
    \begin{tabular}{lcccccc}
    \toprule
     & KCS & Optimised in Fig.~\ref{shipgan_f17} & KVLCC & Optimised in Fig. \ref{shipgan_f24} & Crew Supply & Optimised in Fig.~\ref{shipgan_f25}\\
     \midrule
     $L_{lw}$ & 232.5 & 229.6 & 325.5 & 320.7 & 34.7 & 34.7\\
    $B_{wl}$ & 32.2 & 31.8 & 58 & 58 & 6 & 5.8\\
    $T$ & 10.8 & 10.5 & 20.8 & 20.8 & 0.9 & 0.9\\
   $\nabla$ & 53811 & 51370 & 314446 & 301852 & 56.8 & 55.4\\
$C_w$ & 2.48E-03 & 5.93E-04 & 6.81E-03 & 2.65E-03 & 2.66E-03 & 1.03E-03\\
    \bottomrule
    \end{tabular}
    \label{table_2}
    \end{table}

\subsubsection{Conventional optimisation}
We now proceed with another example aiming to test the performance of ShipHullGAN in the context of conventional parametric modelling, where parametric modellers are developed using a specific hull type in order to produce a design space capable of creating design variants around a given parent hull. For this purpose, we assume the crew supply vessel hull shown in Fig.~\ref{shipgan_f2} as the parent hull and extract its closest design, $\mathbf{z}_{cs}$, from the employed design space $\mathcal{Z}$. In sequel, we use $\mathbf{z}_{cs}$ as the parent hull and consider a subspace, $\mathcal{Z}_{{cs}}$, in the neighbourhood of $\mathbf{z}_{cs}$, by appropriately limiting the original design space $\mathcal{Z}$. This is aligned with the conventional parametric modelling approach, in which slight variations of the parent hull are considered. Specifically, this subspace is defined in the range $[0.90\mathbf{z}_{cs},1.10\mathbf{z}_{cs}]$, which permits a 10\% variation from the parent design. The optimisation process in Eq.~\eqref{shipganOptEq3} is executed utilising the newly established subspace. 

  \begin{equation}\label{shipganOptEq3}
    \begin{aligned}
    \textrm{Find } \mathbf{z}^*_{cs}\in\mathbb{R}^{20} \quad & \textrm{such that} \\
    C_w(\mathbf{z}^*_{cs}) = &  \min_{\mathbf{z}_{cs} \in \mathcal{Z}_{{cs}}} C_w(\mathbf{z}_{cs}) \\
    \textrm{subject to} \quad  & 53.96m^3 \leq \nabla \leq 59.64 m^3,\\
    \quad & 33.0\leq L_{wl} \leq 36.4m,\\
    \quad  & 5.55m \leq B_{wl} \leq 6.13m,\\
    \quad & 0.86 \leq T \leq 0.95m.
     \end{aligned}
    \end{equation}

\begin{figure}[htb!]
    \centering
    \includegraphics[width=01\textwidth]{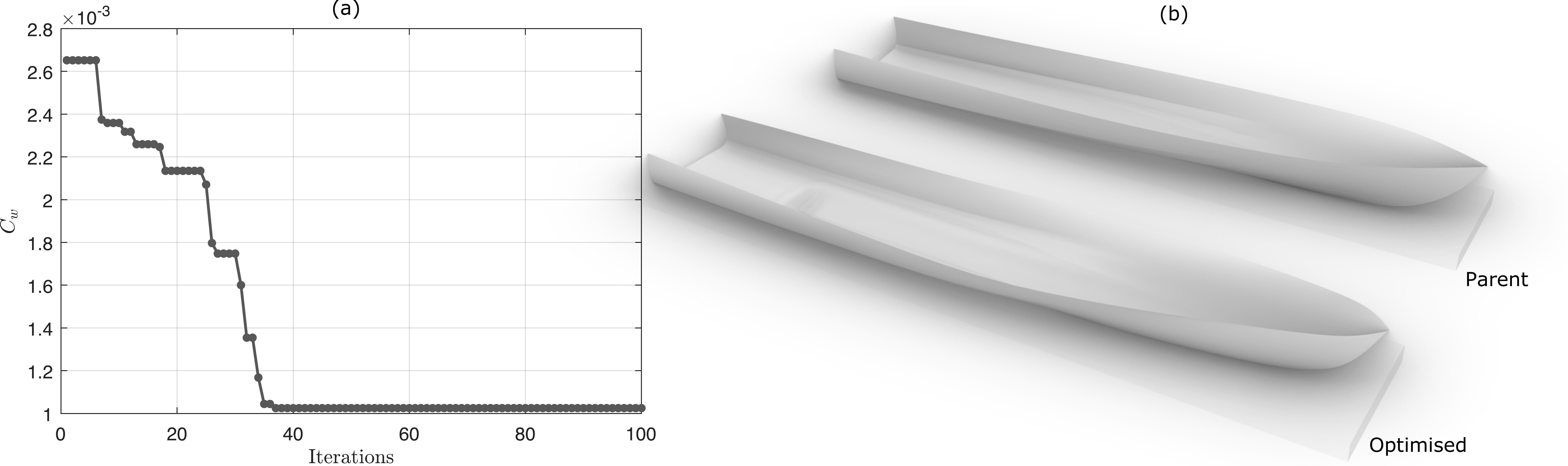}
    \caption{(a) Convergence plots of $C_w$ during the first 100 optimisation iterations performed in $\mathcal{Z}_{{cs}}$. (b) 3D surfaces of $\mathbf{z}_{cs}$ and its optimised variant found in $\mathcal{Z}_{{cs}}$.}
    \label{shipgan_f25}
    \end{figure}

\noindent The results of this experiment are shown in Fig.~\ref{shipgan_f25} while the $C_w$ values of $\mathbf{z}_{cs}$ and its optimised version are $2.66E-03$ and $1.03E-03$, respectively, which show a substantial reduction in the $C_w$. 
Table~\ref{table_2} summarises the results obtained for all three examples of this section.

\section{Concluding remarks}
In this work, we demonstrated the first application of deep convolutional generative adversarial networks for the parametric modelling and design optimisation of ship hulls. We first present a new architecture of GANs employing a space-filling layer to ensure the generator's capacity to cover all design classes in the design space. We have additionally introduced geometric moments (GMs) to the network model, along with an appropriate shape representation in the form of a Shape Signature Tensor (SST). GMs provide rich information about the overall design's geometric structure, and specifically for the ship design case, they also induce the notion of physics. This approach results in a robustly trained generator consistently producing geometrically valid design instances and practically feasible hull form shapes. The capability of the developed ShipHullGAN model is assessed using a variety of metrics and demonstrated with the help of a series of indicative ship hull design optimisation problems. 

\subsection{Future work}
Future extensions of this work aim to conduct a large-scale and multiobjective shape optimisation with the integration of all the components of resistance evaluated with CFD solvers. Moreover, we aim to target the enforcement of the physics-informed component by training ShipHullGAN simultaneously for physics (similar to reduce-order modelling) and geometries, with a fully connected layer for physics prediction. 

Furthermore, we also plan to investigate the potential benefits of incorporating harmonic mapping \cite{shi2008harmonic} to determine whether it can enhance the representation and reconstruction of shapes in the training dataset, and subsequently improve the design generation capabilities of the proposed ShipHullGAN model.

\section*{Acknowledgements}
We would like to acknowledge Professor Grigorios Grigoropoulos, School of Naval Architecture \& Marine Engineering, National Technical University of Athens (NTUA), Greece, and Dr Matteo Diez and Dr Andrea Serani for their support in providing some of the design geometries used in the training dataset. This work received funding from: 
\begin{enumerate}
    \item the Royal Society under the HINGE (Human InteractioN supported Generative modEls for creative designs) project via their \say{International Exchanges 2021 Round 2} funding call, PI: P.D. Kaklis, Co-PI: K.G. Lambert,
    \item the European Union's Horizon 2020 research and innovation programme under the Marie Skłodowska-Curie grant agreement No 860843, PI for the University of Strathclyde: P.D. Kaklis, and
    \item the Nazarbayev University, Kazakhstan under the grant: \say{SOFFA – PHYS: Shape Optimisation of Free-form Functional surfaces using isogeometric Analysis and Physics-Informed Surrogate Models}(grant award No. 11022021FD2927), PI:\ K.V. Kostas.
\end{enumerate}
\bibliographystyle{elsarticle-num}
\bibliography{refs}
    
\end{document}